\documentclass[twoside]{article}
\usepackage[utf8]{inputenc}
\usepackage{amsmath}
\usepackage{amsfonts}
\usepackage{amssymb}
\usepackage{graphicx}
\usepackage{color}
\usepackage[normalem]{ulem}
\usepackage{bbm, relsize}
\usepackage[left=2cm,right=2cm,top=2cm,bottom=2cm]{geometry}
\usepackage{algorithm,algorithmic,multirow}
\newenvironment{Proof}{\noindent{\sc Proof.}}{\qed}
\newtheorem{theorem}{Theorem}[section]
\newtheorem{lemma}{Lemma}[section]
\newtheorem{cor}{Corollary}[section]
\newtheorem{rem}{Remark}[section]
\newtheorem{definition}{Definition}[section]
\newtheorem{prop}{Proposition}[section]

\newcommand{\qed}{\hfill$\Box$\par\medskip}

\def\bhag#1{\noindent
\setcounter{equation}{0}
\section{#1}
}

\def\RR{{\mathbb R}}
\def\CC{{\mathbb C}}
\def\ZZ{{\mathbb Z}}

\def\PPI{{{\rm I}\kern-1pt\Pi}}

\def\b #1;{{\bf #1}}
\def\x{{\bf x}}
\def\k{{\bf k}}
\def\y{{\bf y}}
\def\u{\mathbf{u}}

\def\z{{\bf z}}

\def\O{{\cal O}}

\def\C{{\mathcal C}}

\def\EE{{\mathbb E}}

\def\argmax{\mathop{\hbox{{\rm arg max}}}}

\def\be{\begin{equation}}
\def\ee{\end{equation}}
\def\bea{\begin{eqnarray}}
\def\eea{\end{eqnarray}}
\def\eref#1{(\ref{#1})}
\def\disp{\displaystyle}

\def\donchitre#1#2{\vskip 6.5cm\noindent
\parbox[t]{1in}{\special{eps:#1.eps x=6.5cm y=5.5cm}}
\hbox to 7cm{}\parbox[t]{0.0cm}{\special{eps:#2.eps x=6.5cm y=5.5cm}}}

\def\XX{{\mathbb X}}

\def\bs#1{{\boldsymbol{#1}}}

\title{A   witness function based construction of discriminative models using Hermite polynomials}
\author{
 H.~N.~Mhaskar\thanks{
Institute of Mathematical Sciences, Claremont Graduate University, Claremont, CA 91711, U.S.A.. The research of this author is supported in part   by the Office of the Director of National Intelligence (ODNI), Intelligence Advanced Research Projects Activity (IARPA), via 2018-18032000002.
\textsf{email:} hrushikesh.mhaskar@cgu.edu} 
 \ , A.~Cloninger\thanks{Department of Mathematics, University of California San Diego, San Diego, CA 92093, U.S.A..  The research of this author is supported in part by the National Science Foundation project DMS-1818945.
 \textsf{email:} acloninger@ucsd.edu}
 \ , and X.~Cheng\thanks{Department of Mathematics, Duke University, Durham, NC, U.S.A..  The research of this author is supported in part by the National Science Foundation project DMS-1818945.
 \textsf{email:} xiuyuan.cheng@duke.edu}
 }

\begin{document}
\maketitle
\begin{abstract}
In machine learning, we are given a dataset of the form $\{(\x_j,y_j)\}_{j=1}^M$, drawn as  i.i.d. samples from an unknown probability distribution $\mu$; the marginal distribution for the $\x_j$'s being $\mu^*$. 
We propose that rather than using a positive kernel such as the Gaussian for estimation of these measures, using a non-positive kernel that preserves a large number of moments of these measures yields an optimal approximation.
We use multi-variate Hermite polynomials for this purpose, and prove optimal and local approximation results in a supremum norm in a probabilistic sense.
Together with a permutation test developed with the same kernel, we prove that the kernel estimator serves as a ``witness function'' in classification problems. 
Thus, if the value of this estimator at a point $\x$ exceeds a certain threshold, then the point is reliably in a certain class.  
This approach can be used to modify pretrained algorithms, such as neural networks or nonlinear dimension reduction techniques, to identify in-class vs out-of-class regions for the purposes of generative models, classification uncertainty, or finding robust centroids.
This fact is demonstrated in a number of real world data sets including MNIST, CIFAR10, Science News documents, and LaLonde data sets.
\end{abstract}



\bhag{Introduction}
A central problem in machine learning is the following. 
We are given data of the form $\{(\x_j,y_j)\}_{j=1}^M$, where each $\x_j$ is typically in a Euclidean space and $y_j$ is a label associated with $\x_j$. 
The data is assumed to be sampled independently from an unknown probability distribution $\mu$. 
The problem is to learn either the \textbf{generative model} $\mu$ or a functional model for the unknown function $\mathbb{E}_\mu(y|\x)$. 
A problem germane to both of these interpretations is to learn the generative model for the $\x$-component; i.e., the marginal distribution $\mu^*$ of $\x$.

In the context of classification problems, the labels $y_j$ are limited to a finite set, which may be identified with $\{1,\ldots, m\}$, where $m$ is the number of classes.
Corresponding to each of these classes, say the $j$-th class, one can define a probability distribution $\mu_j^*(\x)$ giving the probability that the label $j$ is associated with $\x$. These measures can also be thought of as the \textbf{discriminative models} $\mu(j|\x)$.
The classification problem is then equivalent to the approximation of the function, 
$$
\x\mapsto\argmax_{1\le j\le m}\mu_j^*(\x).
$$
We have proposed in earlier papers \cite{multilayer, mauropap} a variant of this problem where one seeks to approximate the function 
$$
\x\mapsto\argmax_{1\le j\le m}\chi_j(\x),
$$
where $\chi_j$ is $1$ if the label associated with $\x$ is $j$, and $0$ otherwise. 
The functions $\chi_j$ are manifestly not continuous. 
Nevertheless, using diffusion geometry (see \cite{achaspissue} for an early introduction), we can assume that the feature space is selected judiciously to be a data-defined manifold on which the supports of these functions are well separated; at least, the set of discontinuities of $\chi_j$ is not too large. 
The use of localized kernels developed in \cite{mauropap, tauberian} on the unknown manifold  is then expected to yield a good approximation to the $\chi_j$'s and hence, a good classification.
A theory for this sort of approximation is well developed in many papers, e.g., \cite{fasttour, mauropap, modlpmz, eignet}, illustrated with some toy examples in \cite{mauropap, compbio}, and in connection with diabetic blood sugar prediction in a clinical data set in \cite{mhas_sergei_maryke_diabetes2017}.
A bottleneck in this approach is the construction of the right eigensystem to approximate with. 
Another drawback of this strategy is the lack of out-of-sample extension; i.e., the model depends upon the data set at hand, the entire computation needs to be repeated with the appearance of any new data.

In this paper, we seek to approximate more directly the measures $\mu_j^*$. 
Unlike the approach in \cite{mauropap}, this needs to be done without knowing the value of $\mu_j^*$ even at training data points. 
These values are just not available. 
Further,
we do not make any assumptions on the structure of the support of these measures. 
Rather than using a data dependent orthogonal system,  we use the localized  kernels developed in \cite{mohapatrapap, hermite_recovery}, based on Hermite polynomials. 
This has several advantages.
\begin{enumerate}
\item The kernels are well defined on the entire Euclidean space and their theoretical properties are well investigated.
Therefore, the use of these kernels obviates the problem of out-of-sample extension. 
Indeed, we use this kernel to generate labels for new, unseen possible points in the feature space. 
\item Although the use of these kernels does involve some tunable parameters, the constructions are more robust with respect to the choice of these parameters, compared with the role of the parameters in the diffusion geometry setting.
\item In view of the Mehler identity, these kernels can be computed efficiently even in high dimensions using only the inner products of the data points (cf. Section~\ref{kernimpsect}).
\item It is shown in \cite{convtheo, chuigaussian} that these kernels can be implemented as Gaussian networks with arbitrary accuracy. 
\end{enumerate}

Since the values of $\mu_j^*$ are not known, we resort to a straightforward kernel estimation. 
In classical probability estimation, one typically uses a positive kernel to ensure that the estimate is also a probability measure. 
From an approximation theory point of view, this is doomed to give a poor approximation.
Instead, the use of our kernels, which are not  positive, ensures that increasingly many moments of the approximation are the same as those of the original measure.
This ensures an optimal approximation, albeit not by probability measures.
 Assuming absolute continuity of $\mu_j^*$ with respect to $\mu^*$ and that of $\mu^*$ with respect to the Lebesgue measure on the ambient space, we obtain \textbf{local} rates of approximation in terms of the number of samples in the training set and the smoothness of the Radon-Nikodym derivatives of these measures.
 We note that the assumption that $\mu^*$ is absolutely continuous with respect to the Lebesgue measure on the ambient space is restrictive, and further research is required to examine how to relax this condition. 
 Nevertheless, this assumption seems reasonable in the present context in order to take into account noise in the data.
In contrast to classical machine learning, our estimates are pointwise in a probabilistic sense rather than in the sense of an $L^2$ norm.
A technical novelty in this analysis is a Videnskii-type inequality that estimates the derivative of a weighted polynomial on a cube in terms of the norm of the weighted polynomial on the \textbf{same} cube, the estimates being independent of the center of this cube. 

In one-hot classification, we need to estimate $\mu_j^*$ as if there are two classes: labeled $1$ if the actual class label is $j$ and $-1$ otherwise; i.e., we treat two measures at a time, $\mu_j^*$ and $\sum_{k\not=j}\mu_j^*$. In general, the question of detecting where two distributions deviate, and whether that deviation is significant, is of great interest both in machine learning and in various scientific disciplines.  
We refer to the difference between the estimators of these two measures as a \textbf{witness function}.
 The approach of building a kernel based witness function has been developed by Gretton et al \cite{gretton2012kernel}.  
 In these works, the witness function that identifies where two samples deviate comes as a biproduct of determining the maximum mean discrepancy (MMD) between the two distributions and create a measure of global deviation between the distributions.  
The paper \cite{cheng2017two} describes how the power of the global test is affected by changing the definition of locality in the kernel. 
 In the present work, we examine the local deviation of distributions as measured by the witness function, and provide theory and algorithms for determining whether the deviations are significant locally.   
 This is important for a number of applications, where it is important to highlight why two distributions deviate, and determine whether specific bumps in the witness function are a product of structure or of noise. 
  Each experiment in Section \ref{exptsect} relies on identifying these local deviations.
 
In Section \ref{toycirclesect}, we demonstrate experimentally that introducing the Hermite kernel significantly increases the power of detecting local deviations compared to the Gaussian kernel traditionally used in MMD.

An important application of determining the local deviation between distributions is in accurately determining the confidence of predictions or generated points using neural networks.   
There have been many reported cases of decisions being made by neural networks for points that lie far away from the training data, and yet are assigned high confidence predictive value \cite{guo2017calibration, hendrycks2016baseline}.   
Moreover, it has been recently shown that this is an inherent property of ReLU networks \cite{hein2018confidence}.  
While there have been a number of attempts to alleviate the issue for predictive nets, there hasn't been nearly as much work for determining out-of-distribution regions for generative networks and Variational Autoenconders, where sampling from out-of-distribution regions leads to non-natural images or blurring between multiple classes of images.  
We discuss avoiding out-of-distribution and out-of-class sampling of Variational Autoencoders in Section \ref{mnistvaesect}.  
The use of a witness function for model comparison of GANs is studied in \cite{sutherland2016generative}. However, that paper only considers the raw size of the witness function without establishing a local baseline, and does not provide a theory of how the empirical witness function deviates from the true witness function.  Most approaches to mitigating out-of-distribution high confidence prediction require changing the training objective for the network.  We instead examine a pretrained VGG-16 network and determine a method for detecting outliers and likely misclassified points in Section \ref{CNNsect}.   

Certain scientific applications also benefit from recognition of the where two distributions deviate.  The topic of clustering documents based off of a co-occurance of words has been a topic of significant consideration \cite{shahnaz2006document,wang2009multiscale}.  However, most approaches based on k-means in an embedding space can be biased by outliers and clusters with small separation.  We examine the uses of the witness function for determining in class vs out-of-training distribution points in a term document embedding in Section \ref{SNembeddingsect} using the Hermite kernel, which exhibits better edges between close or overlapping clusters.  Another application is in propensity matching, in which the problem is to identify bias in which groups received or didn't receive treatment in an observational trial.  Propensity matching boils down to modeling the probability of being in one class vs the other, traditionally with a logistic regression, and using such probability for subsequent matching of treated and untreated patients \cite{rosenbaum1983central}.  The uses of propensity matching in literature are too numerous to cite here, but we refer readers to the following article outlining both the importance and its drawbacks \cite{king2016propensity}.  We instead consider a distance based matching using the nonlinear Hermite kernel in Section \ref{lalondesect}, and demonstrate that viewing this problem as finding local deviations of the witness function allows for the benefits of both an unsupervised distance based algorithm like proposed in \cite{king2016propensity} and a simple 1D similar to a propensity score that describes the bias in treatment.

To summarize, we illustrate our theory using the following examples:
\begin{enumerate}
\item\label{toy} A toy example of detecting the differences, with low false discovery rate, in support of a measure supported on a circle verse one supported on an ellipse (Section~\ref{toycirclesect}),
\item\label{digit} Discovering the boundaries between  classes in the latent space of a variational autoencoder, and generating ``prototypical'' class examples in the MNIST data set (Section~\ref{mnistvaesect}),
\item\label{cifar10} Prospectively quantifying the uncertainty in classification of the VGG-16 network trained on CIFAR10 (Section~\ref{CNNsect}),
\item\label{documents} Determining robust cluster centroids in a document-term matrix of Science News documents (Section~\ref{SNembeddingsect}), and
\item\label{social} Discovering the propensity of treatment for people in the LaLonde job training data set (Section~\ref{lalondesect}).

\end{enumerate}
%

We develop the necessary background and notation for Hermite polynomials and associated kernels and other results from the theory of weighted polynomial approximation in Section~\ref{backsect}. 
Our main theorems are discussed in Section~\ref{recoverysect}, and proved in Section~\ref{pfsect}.
The algorithms to implement these theorems are given in Section~\ref{permutationsect}, and the results of their application in different experiments are given in Section~\ref{exptsect}.

\bhag{Background and notation}\label{backsect}
\subsection{Hermite polynomials}\label{hermitesect}
A good preliminary source of many identities regarding Hermite polynomials is the book \cite{szego} of Szeg\"o or the Bateman manuscript \cite{batemanvol2}.

The orthonormalized Hermite polynomial $h_k$ of degree $k$ is defined by the Rodrigues' formula:
\be\label{rodrigues}
h_k(x)=\frac{(-1)^k}{\pi^{1/4}2^{k/2}\sqrt{k!}}\exp(x^2)\left(\frac{d}{dx}\right)^k\exp(-x^2).
\ee
Writing $\psi_k(x)=h_k(x)\exp(-x^2/2)$, one has the orthogonality relation for $k, j\in \ZZ_+$,
\be\label{uniortho}
\int_\RR \psi_k(x)\psi_j(x)dx=\left\{\begin{array}{ll}
1, & \mbox{if $k=j$, }\\
0, & \mbox{if $k\not=j$.}
\end{array}\right.
\ee

In multivariate case, we adopt the notation $\x=(x_1,\cdots, x_q)$. The orthonormalized Hermite function is defined by
\be\label{multihermite}
\psi_\k(\x)=\prod_{j=1}^q\psi_{k_j}(x_j).
\ee
In general, when univariate notation is used in multivariate context, it is to be understood in the tensor product sense as above; e.g., $\k!=\prod_{j=1}^q k_j!$,
$\x^\k=\prod_{j=1}^qx_j^{k_j}$, etc. The notation $|\cdot|_p$ will denote the Euclidean $\ell^p$ norm, with the subscript omitted when $p=2$. 

We define 
\be\label{projdef}
\mathsf{Proj}_m(\x,\y)=\sum_{|\k|_1=m}\psi_\k(\x)\psi_\k(\y)
\ee

Let $H: [0,\infty)\to [0,1]$ be a $C^\infty$ function, $H(t)=1$ if $t\in [0,1/2]$, $H(t)=0$ if $t\ge 1$. We define
\be\label{summkerndef}
\Phi_n(H;\x,\y)=\Phi_n(\x,\y)=\sum_{\k\in\ZZ_+^q}H\left(\frac{\sqrt{|\k|_1}}{n}\right)\psi_\k(\x)\psi_\k(\y)=\sum_{m=0}^\infty H\left(\frac{\sqrt{m}}{n}\right)\mathsf{Proj}_m(\x,\y).
\ee

\noindent\textbf{Constant convention:}\\

In the sequel, $c, c_1,\cdots$ will denote positive constants depending upon $q$, $H$, and other fixed quantities in the discussion, such as the norm. Their values
may be different at different occurrences, even within a single formula. \qed

The following lemma \cite[Lemma~4.1]{hermite_recovery} lists some important properties of $\Phi_n$ (the notation in \cite{hermite_recovery} is somewhat different; the kernel $\Phi_n$ above is $n^q\Phi_n$ in the notation of \cite{hermite_recovery}).

\begin{lemma}\label{kernlemma} 
Let $S>q$ be an integer.  \\
{\rm (a)} For $\x,\y\in\RR^q$, $n=1,2,\cdots$,
\be\label{hermite_localization}
|\Phi_n(\x,\y)| \le \frac{cn^q}{\max(1,(n|\x-\y|)^S)}.
\ee
In particular,
\be\label{hermite_max_norm}
|\Phi_n(\x,\y)| \le cn^q.
\ee
{\rm (b)} For $\x\in\RR^q$, $n=1,2,\cdots$, $1\le p < \infty$
\be\label{phinintest}
\int_{\RR^q}|\Phi_n(\x,\y)|^pd\y \le cn^{q(1-1/p)}.
\ee
\end{lemma}

\subsection{Weighted polynomials}\label{wtpolysect}

For $n>0$ (not necessarily an integer), let $\Pi_n^q=\mathsf{span}\{\psi_\k : |\k|_1<n^2\}$. An element of $\Pi_n^q$ has the form $P(\x)\exp(-|\x|^2/2)$ for a polynomial $P$ of total degree $n^2$.
The following proposition lists a few important properties of these spaces (cf. \cite{mhasbk, gaussbern, mohapatrapap}).
\begin{prop}\label{wtpolyprop}
Let $n >0$, $P\in\Pi_n^q$, $1\le p\le\infty$. \\
{\rm (a) (Infinite-finite range inequality)} For any $\delta>0$, there exists $c=c(\delta,p)$ such that
\be\label{rangeineq}
\|P\|_{p,\RR^q\setminus [-\sqrt{2}n(1+\delta), \sqrt{2}n(1+\delta)]^q} \le c_1e^{-cn^2}\|P\|_{p,[-\sqrt{2}n(1+\delta), \sqrt{2}n(1+\delta)]^q}
\ee
{\rm (b) (MRS identity)} We have
\be\label{mrsidentity}
\|P\|_\infty=\|P\|_{\infty,[-\sqrt{2}n, \sqrt{2}n]^q}.
\ee
{\rm (c) (Bernstein inequality)} There is a positive constant $B$ depending only on $q$ such that
\be\label{bernineq}
\||\nabla P|\|_p \le Bn\|P\|_p.
\ee
\end{prop}

In view of Proposition~\ref{wtpolyprop}, we refer to the cube $[-\sqrt{2}n, \sqrt{2}n]^q$ as the critical cube. When $q=1$, it is often called the MRS (Mhaskar-Rakhmanov-Saff) interval.
The following corollary is easy to prove using Proposition~\ref{wtpolyprop}, parts (b) and (c).
\begin{cor}\label{coveringcor}
Let $n>0$, $\C\subset I_{n,q}=[-\sqrt{2}n, \sqrt{2}n]^q$ be a finite set satisfying
\be\label{coveringmeshnorm}
\max_{\x\in I_{n,q}}\min_{\y\in\C}\|\x-\y\| \le 1/(2Bn).
\ee
Then for any $P\in\Pi_n^q$,
\be\label{finitemaxnorm}
\max_{\y\in\C}|P(\y)| \le \|P\|_\infty \le 2\max_{\y\in\C}|P(\y)|.
\ee
There exists a set $\C$ as above with $|\C|\sim n^{2q}$.
\end{cor} 

If $r>0$, $\x_0\in\RR^q$, we denote the ball of radius $r$ around $\x_0$ by
\be\label{balldef}
\mathbb{B}(\x_0,r)=\{\y\in\RR^q : \|\x_0-\y\|_\infty \le r\}.
\ee

\subsection{Function approximation}\label{approxsect}
In this section, we describe some results on approximation of functions.
If $1\le p\le \infty$, $f\in L^p$, 
\be\label{degapprox}
E_{n,p}(f)=\min_{P\in\Pi_n^q}\|f-P\|_p.
\ee
The symbol $X^p$ denotes the set of all $f\in L^p$ for which $E_{n,p}(f)\to 0$ as $n\to\infty$. Thus, $X^p=L^p$ if $1\le p<\infty$, and $X^\infty=C_0$. For $\gamma>0$, the smoothness class $W_{p,\gamma}$ comprises $f\in X^p$ for which
\be\label{sobolevnormdef}
\|f\|_{W_{p,\gamma}}=\|f\|_p+\sup_{n\ge 0}2^{n\gamma}E_{2^n,p}(f) <\infty.
\ee
In \cite{mhasbk, tenswt}, we have given a characterization of the spaces $W_{p,\gamma}$ in terms of the constructive properties of $f$ in terms of divided differences and the bounds near $\infty$. 
We define
\be\label{sigmadef}
\sigma_n(f)(\x)=\int_{\RR^q}\Phi_n(\x,\y)f(\y)d\y, \qquad f\in L^1+L^\infty,\ n>0, \ \x\in\RR^q.
\ee
The following proposition is routine to prove using Lemma~\ref{kernlemma}(b) with $p=1$:
\begin{prop}\label{goodapproxprop}
{\rm (a)} If $n>0$ and $P\in \Pi_{n/\sqrt{2}}$, then $\sigma_n(P)=P$.\\
{\rm (b)} If $f\in L^p$, $n>0$, then
\be\label{sigmaopbd}
\|\sigma_n(f)\|_p\le c\|f\|_p, \quad E_{n,p}(f)\le \|f-\sigma_n(f)\|_p\le cE_{n/\sqrt{2},p}(f).
\ee
{\rm (c)} We have
\be\label{sobolchar}
\|f\|_{W_{p,\gamma}}\sim \|f\|_p+\sup_{n\ge 0}2^{n\gamma}\|f-\sigma_{2^n}(f)\|_p \sim \sup_{n\ge 0}2^{n\gamma}\|\sigma_{2^{n-1}}(f)-\sigma_{2^n}(f)\|_p.
\ee
\end{prop}

Next, we describe  local smoothness classes. 
If $\x_0\in\RR^q$, $\gamma>0$ the local smoothness class $W_{p,\gamma}(\x_0)$ comprises functions $f\in L^p$ with the following property: There exists a neighborhood $U$ of $\x_0$ such that for every $C^\infty$ function $\phi$ supported on $U$, $\phi f\in W_{p,\gamma}$. 
We note that the quantity $\gamma$ is expected to depend upon $\x_0$.
The following characterization of local smoothness classes can be obtained by imitating arguments in \cite{mohapatrapap}.

\begin{theorem}\label{localchartheo}
Let $1\le p\le \infty$, $f\in X^p$, $\gamma>0$, $\x_0\in\RR^q$. The following statements are equivalent:\\
{\rm (a)} $f\in W_{p,\gamma}(\x_0)$.\\
{\rm (b)} There exists $r=r(f,\x_0,p,\gamma)>0$  such that
$$
\sup_{n\ge 0}n^{\gamma}\|f-\sigma_n(f)\|_{p, \mathbb{B}(\x_0,r)}<\infty.
$$
{\rm (c)} There exists $r_1=r_1(f,\x_0,p,\gamma)>0$ such that
$$
\sup_{n\ge 0}2^{n\gamma}\|\sigma_{2^{n-1}}(f)-\sigma_{2^n}(f)\|_{p, \mathbb{B}(\x_0,r_1)}<\infty.
$$
\end{theorem}
If $\x_0\in\RR^q$, $1\le p\le\infty$, $f\in W_{p,\gamma}(\x_0)$, and 
part (b) of Theorem~\ref{localchartheo} holds with $\mathbb{B}(\x_0,r)$ for some $r>0$, we define
\be\label{locsobnorm}
\|f\|_{p,\gamma,\x_0,r}=\|f\|_p+\sup_{n\ge 0}n^{\gamma}\|f-\sigma_n(f)\|_{p, \mathbb{B}(\x_0,r)},
\ee
where we note that this defines a norm since we have used the norm of $f$ on the entire $\RR^q$ as one of the summands above.

\bhag{Recovery of measures}\label{recoverysect}
In the two sample problem, one has samples $\C_j$ from  distributions $\mu_j$, $j=1,2$, and associates the label $1$ with $C_1$, $-1$ with $\C_2$. 
The task of a witness function is to determine if in the neighborhood of a given point $\mu_1$ dominates $\mu_2$ or the other way round, or if they are equal. 
If both the distributions are absolutely continuous with respect to the Lebesgue measure on $\RR^q$ with smooth densities $f_1$, $f_2$ respectively, then 
Theorem~\ref{localchartheo} suggests that $\sigma_n(f_1-f_2)$, or its Monte-Carlo estimator using the samples should work as a witness function. 
If the Lebesgue measure were a probability distribution on $\RR^q$, then it would be easy to put probabilistic bounds to make this more precise.
Since this is not the case, we take the viewpoint that $\C_1\cup \C_2$ is a sample from a ground probability distribution $\mu^*$ with smooth density $f$, and $F$ is another smooth function that takes approximately the value $1$ on $\C_1$, $-1$ on $\C_j$. 
Then  a candidate for the witness function is given by
$$
\sigma_n(Ff)(\x)=\int_{\RR^q}F(\y)f(\y)\Phi_n(\x,\y)d\y=\int_{\RR^q} F(\y)\Phi_n(\x,\y)d\mu^*(\y).
$$
With this re-formulation, we no longer need to restrict ourselves to two classes, $F$ can be chosen to approximate any number of class values, or can even be just any smooth function.
The following theorem makes these sentiments more precise in terms of the Monte-Carlo approximation to the last integral above.

Next, we discuss the  robustness of this witness function.
For this purpose, we assume noisy data of the form $(\y,\epsilon)$, with a joint probability distribution $\tau$ and with $\mu^*$ being the marginal distribution of $\y$ with respect to $\tau$. 
In place of $F(\y)$, we consider a noisy variant $\mathcal{F}(\y,\epsilon)$, and denote
\be\label{Fdef} 
F(\y)=\mathbb{E}_\tau(\mathcal{F}(\y,\epsilon)|\y).
\ee
It is easy to verify using Fubini's theorem that if $\mathcal{F}$ is integrable with respect to $\tau$ then
for any $\x\in\RR^q$,
\be\label{iterexp}
\sigma_n(Ff)(\x)=\mathbb{E}_\tau(\mathcal{F}(\y,\epsilon)\Phi_n(\x,\y)).
\ee

A part of our theorem below uses the Lambert function
defined by 
\be\label{lambertfn}
\mathcal{W}(ze^z)=z, \quad W(z)>1 \mbox{ if $z\ge 1$}.
\ee
It is known that
\be\label{lambertass}
\mathcal{W}(x)=\log x-\log\log x +o(1), \quad x\to\infty.
\ee

\begin{theorem}\label{locprobesttheo}
Let $\tau$ be a probability distribution on $\RR^q\times \Omega$ for some sample space $\Omega$, $\mu^*$ be the marginal distribution of $\tau$ restricted to $\RR^q$.
We assume that $\mu^*$ is absolutely continuous with respect to the Lebesgue measure on $\RR^q$ and denote its density by $f$.
Let $\mathcal{F} : \RR^q\times \Omega\to \RR$ be a bounded function, and $F$ be defined by \eref{Fdef}. 
Let $\x_0\in\RR^q$, $\gamma>0$,  $\delta>0$,
 $Ff\in W_{\infty,\gamma}(\x_0)$, and $r$ be chosen such that $\|Ff\|_{\infty,\gamma,\x_0,r}<\infty$ (cf. \eref{locsobnorm}). 
Let $M\ge 1$, $Y=\{(\y_1,\epsilon_1),\cdots,(y_M,\epsilon_M)\}$ be a set of random samples chosen i.i.d. from $\tau$, and define
\be\label{festimator}
\widehat{F}(\x)=\widehat{F}(Y;\x)=\frac{1}{M}\sum_{j=1}^M \mathcal{F}(\y_j,\epsilon_j) \Phi_n(\x,\y_j), \qquad \x\in\RR^q.
\ee
 Then there exists $c_1>0$ such that for every $n\ge 1$ and $r\ge c_1/n^2$,
\be\label{masterprobest}
\mathsf{Prob}_{\tau}\left(\left\|\widehat{F}(Y;\circ)-Ff\right\|_{\infty,\mathbb{B}(\x_0,r)}\ge  c_2\|\mathcal{F}\|_\infty n^q
\sqrt{\frac{\log(c_3r^{q}\exp(q/r)n^{5q}/\delta)}{M}}+n^{-\gamma}\|Ff\|_{\infty,\gamma,\x_0,r}\right)\le \delta(r/n)^q.
\ee
In particular, writing
$B=c_3r^q\exp(q/r)/\delta$,  $\Gamma=(2\gamma+2q)/(5q)$, and
\be\label{nmdef}
n=C_1B^{-1/(5q)}\exp\left(\frac{1}{2q+2\gamma}\mathcal{W}\left(\Gamma B^\Gamma M\right)\right)\sim \left(\frac{M}{\log M}\right)^{1/(2q+2\gamma)},
\ee
we obtain for $r\ge c_1/n^2$ that
\be\label{locprobest}
\mathsf{Prob}_{\tau}\left(\left\|\widehat{F}(Y;\circ) -Ff\right\|_{\infty,\mathbb{B}(\x_0,r)}\ge c_4\frac{\|\mathcal{F}\|_\infty+\|Ff\|_{\infty,\gamma,\x_0,r}}{n^\gamma}\right)\le \delta(r/n)^q.
\ee
\end{theorem}

\begin{rem}\label{zerorem}
{\rm
In the case when $Ff\in C^\infty(\mathbb{B}(\x_0,r))$, (in particular, when $Ff\equiv 0$ on $\mathbb{B}(\x_0,r)$), one may choose $\gamma$ to be arbitrarily large, although the constants involved may depend upon the choice of $\gamma$.
\qed}
\end{rem}
\begin{rem}\label{globalrem}
{\rm
We note that  the critical cube $[-\sqrt{2}n, \sqrt{2}n]^q$ can be partitioned into $\sim (n/r)^q$ subcubes of radius $r$. On each of these subcubes, say the subcube with center $\x_0$  the function $Ff$ is in a different smoothness class $\gamma(\x_0)$. Therefore, Theorem~\ref{locprobesttheo} implies an estimate for the entire critical cube with probability at least $1-\delta$.
\qed}
\end{rem}

\begin{rem}\label{gendiscrem}
{\rm
Taking $\mathcal{F}\equiv 1$, $\widehat{F}$ approximates the generative model $\mu^*$.  Thus, our construction can be viewed both as an estimator of the generative model as well as a witness function for the discriminative model.
\qed}
\end{rem}
%
%
%
%
%

\vskip -1cm
\bhag{Algorithms}\label{permutationsect}

\subsection{Implementation of the kernels}\label{kernimpsect}
In this sub-section, we describe the construction of the kernels $\Phi_n$. We remark firstly that the univariate Hermite functions can be computed using the recurrence relations 
\bea\label{recurrence}
x\psi_{j-1}(x)&=&\sqrt{\frac{j}{2}}\psi_j(x) + \sqrt{\frac{j-1}{2}}\psi_{j-2}(x),\quad j=2,3,\cdots,\nonumber\\
&&\psi_0(x)=\pi^{-1/4}\exp(-x^2/2),\ \psi_1(x)=\sqrt{2}\pi^{-1/4}x\exp(-x^2/2).
\eea
Next, we recall the Mehler identity, valid for $w\in\CC$, $|w|<1$ and $\x,\y\in\RR^q$:
\be\label{mehler}
 \sum_{\k\in\ZZ^q} \psi_\k(\x)\psi_\k(\y)w^{|\k|_1}= \sum_{m=0}^\infty w^m \mathsf{Proj}_m(\x,\y)=\frac{1}{(\pi(1-w^2))^{q/2}}\exp\left(\frac{4w\x\cdot\y-(1+w^2)(|\x|^2+|\y|^2)}{2(1-w^2)}\right).
\ee
It is clear that the projections $\mathsf{Proj}_m(\x,\y)$ depend only on $\x\cdot\y$, $\x\cdot\x$ and $\y\cdot\y$; in particular, that they are rotation independent. 
Denoting by $\theta$ the acute angle between $\x$ and $\y$, we may thus write
\be\label{reduceproj1}
\mathsf{Proj}_m(\x,\y)=\sum_{j=0}^m \psi_j(|\x|)\psi_j(|\y|\cos\theta)\sum_{\ell=0}^{m-j}\psi_\ell(0)\psi_\ell(|\y|\sin\theta)\sum_{\k\in\ZZ^{q-2}_+, \atop |\k|_1=m-j-\ell}|\psi_\k(\bs{0})|^2.
\ee
Using the Mehler identity \eref{mehler}, we deduce that
$$
\sum_{r=0}^\infty w^{2r}\sum_{|\k|_1=2r}|\psi_\k(\bs{0})|^2=(\pi (1-w^2))^{-(q-2)/2}=\pi^{1-q/2}\sum_{r=0}^\infty  \frac{\Gamma(q/2+r-1)}{\Gamma(q/2-1) r!}w^{2r}.
$$
Hence,
\be\label{reduceproj}
\mathsf{Proj}_m(\x,\y)=\sum_{j=0}^m \psi_j(|\x|)\psi_j(|\y|\cos\theta)\sum_{\ell=0}^{m-j}\psi_\ell(0)\psi_\ell(|\y|\sin\theta)D_{q-2;m-j-\ell},
\ee
where 
\be\label{reduceprojdetails1}
\psi_\ell(0)=\left\{\begin{array}{ll}
\disp \pi^{-1/4}(-1)^{\ell/2}\frac{\sqrt{\ell!}}{2^{\ell/2}(\ell/2)!},&\mbox{if $\ell$ is even},\\[1ex]
0, &\mbox{if $\ell$ is odd},
\end{array}\right.
\ee
and
\be\label{reduceprojdetails2}
D_{q-2;r}=\left\{\begin{array}{ll}
\disp\pi^{1-q/2}\frac{\Gamma(q/2+r/2-1)}{\Gamma(q/2-1) (r/2)!}, &\mbox{if $r$ is even},\\[1ex]
0, &\mbox{if $r$ is odd}.
\end{array}\right.
\ee

\begin{rem}\label{gaussianrem}
{\rm
A completion of squares shows that
\be\label{mehler_to_gauss}
(1+w^2)(|\x|^2+|\u|^2)-4w\x\cdot\u= (1+w^2)\left|\x-
\frac{2w}{1+w^2}\u\right|^2 +\frac{(1-w^2)^2}{1+w^2}|\u|^2.
\ee
With the choice $w=1/\sqrt{3}$, the  Mehler identity \eref{mehler} then shows that
$$
\sum_{m=0}^\infty 3^{-m/2}\mathsf{Proj}_m(\x,\u)= \left(\frac{3}{2\pi}\right)^{q/2}\exp\left(-|\x-\frac{\sqrt{3}}{2}\u)|^2\right)\exp(-|\u|^2/4).
$$
It is now easy to calculate using orthogonality that
\be\label{kernel_to_gaussian}
\Phi_n(\x,\y)=\left(\frac{3}{2\pi}\right)^{q/2}\int_{\RR^q} \exp\left(-|\x-\frac{\sqrt{3}}{2}\y)|^2\right)\exp(-|\y|^2/4)\widetilde{\Phi}_n(\y,\u)d\u,
\ee
where
\be\label{alt_kern_def}
\widetilde{\Phi}_n(\y,\u)=\sum_{m=0}^\infty 3^{m/2}H\left(\frac{\sqrt{m}}{n}\right)\mathsf{Proj}_m(\y,\u).
\ee
A careful discretization of \eref{kernel_to_gaussian} using a Gauss quadrature formula for Hermite weights exact for polynomials of degree $3n^2$ leads to an interpretation of the kernels $\Phi_n$ in terms of Gaussian networks with fixed variance and fixed centers, independent of the data.
We refer to \cite{convtheo, chuigaussian} for the details.

 There is no training required for these networks.
 However, the mathematical results are applicable only to this specific construction. 
Treating the theorem as a pure existence theorem and then trying to find a Gaussian network by traditional training will not work.
\qed}
\end{rem}
\subsection{Algorithms}\label{algsect}
Our numerical experiments in Section~\ref{exptsect} are designed to illustrate the use of $\widehat{F}$ defined in \eref{festimator} as a discriminative model for two classes, arising with probabilities with densities $f_1f$ and $f_2f$ respectively; i.e., with $F=f_1-f_2$. The quantity $\mathcal{F}$ representing the difference between the corresponding class labels is a noisy version of $F$. Intuitively, if $\widehat{F}(\x)$ is larger than a positive threshold in a neighborhood of $\x_0$, then $f_1$ dominates $f_2$ at $\x_0$, and we may take the label for $\x_0$ to be $1$, and similarly if $\widehat{F}(\x)$ is smaller than a negative threshold.  
When $|\widehat{F}(\x)|$ is smaller than the threshold, then there is uncertainty in the correct label for $\x_0$, whether because it belongs both to the supports of $f_1$ and $f_2$ or because $f(\x_0)$ is small, making the point $\x_0$ itself anomalous. 
In theory, Theorem \ref{locprobesttheo} establishes this threshold as $\disp F(\x_0)f(\x_0)- c\frac{\|\mathcal{F}\|_\infty+\|Ff\|_{\infty,\gamma,\x_0,r}}{n^\gamma}$.
However, it is not possible to estimate this threshold based only on the given data.      

For this reason, we introduce the use of the permutation test \cite{pesarin2001multivariate}.  Permutation test is a parameter-free and nonparametric method of testing hypotheses, namely in this case the null hypothesis that  $Ff = 0$ near $\x_0$. Theorem \ref{locprobesttheo}  shows that if the null hypothesis is true then $|\widehat{F}(\x)|$ is smaller than $c\|\mathcal{F}\|_\infty/n^\gamma$with high probability. In turn, it is easy to create a $\mathcal{F}_1$ for which this is true.  We randomly permute the labels of all points $\y_j$ across all classes, reassign these randomized labels $\mathcal{F}_1(\y_j,\epsilon_j)$. Since $\mathcal{F}_1$ represents the class label, this ensures that we know $\|\mathcal{F}_1\|_\infty$ is the same as $\|\mathcal{F}\|_\infty$, but for its expected value $F_1$, $F_1f=0$.  
  Informally, this means we are mixing the distributions of the classes so that when we redraw new collections of points, each collection is being drawn from the same distribution.  The benefit of this test in our context is that we can sample this randomized distribution a large number of times, and use that distribution to estimate $cn^{-\gamma}$.  This threshold we call $T(\x_j)$, as this is the decision threshold associated to the local area around $\x_0$.  If the two classes had equal sampling density around $\y_j$ (i.e. if $Ff=0$), then if we estimated $T(\x_0)$ correctly, Theorem \ref{locprobesttheo} tells us that the probability $\|\widehat{F}\|_{\infty,\mathbb{B}(\x_0,r)}$ exceeds $T(\x_0)$ is  small.  If, on the other hand,  if $\|\widehat{F}\|_{\infty,\mathbb{B}(\x_0,r)} > A\cdot T(\x_0)$ for some constant $A$ associated with $\|Ff\|_{\infty,\gamma,\x_0,r}$ and $\|\mathcal{F}\|_\infty$, then the hypothesis that $Ff=0$ near $\x_0$ can be rejected.

This suggests two parametric choices in our algorithm, estimating $T(\x_0)$ and $A$.  
Estimating $T(\x_0)$ comes down to estimating the threshold that 
$\|Ff\|_{\infty,\gamma,\x_0,r} (\x_0,r)$ exceeds only $\delta (r/n)^q$ fraction of the time.  
Because each random permutation is independent, the probability of failure for any permutation to exceed $cn^{-\gamma}$ over $N$ permutations is $(1 - \delta (r/n)^q)^N$.  
One can choose a desired probability of failure $\alpha$ and compute a number of permutations $N$.  
The statistic $T(\x_0)$ in this case would be the maximum size of the local infinity norm across all permutations.  If we wish to avoid taking the maximum of $N$ random variables, it is also possible to increase the size of $N$ and take the $1 - \delta (r/n)^q$ quantile of the random variables.  

For now, we take $A$ to be a parameter of choice, with the fact that $A\ge 1$, rejection of the $Ff=0$ assumption scales continuously with $A$, and $A$ much larger than $1$ becomes far too restrictive.   Details of the entire algorithm can be found in Algorithm \ref{permutationalgorithm} for the two class test, and Algorithm \ref{permutationalgorithm_multipleclass} for the multiple class test.
 This algorithm returns the empirical witness function $\widehat{F}(Z)$, as well as a decision $D(Z)$ as to whether $\widehat{F}(Z)$ is significantly nonzero (i.e. whether $f_1(\z_i) = f_2(\z_i)$ or $f_1(\z_i) \neq f_2(\z_i)$).   

\begin{algorithm}[h]
\begin{algorithmic}[1]
\item[{\rm a)}] \textbf{Input:} Data sets $X$ and $Y$, points $Z=\{\z_1, ..., \z_K\}$ at which to inspect significance, level of confidence $A$
\item[{\rm b)}] \textbf{Output:} Estimate of $\widehat{F}(\z_j)$ for all $\z_j\in Z$ and estimate of whether $\widehat{F}(\z_j)\neq 0$.
\STATE $\alpha \gets 0.05$
\STATE $\y \gets X\cup Y $
\STATE $M \gets \left\vert X\right\vert + \left\vert Y \right\vert$ 
\STATE $c_j \gets \begin{cases}1, & \textnormal{ if } \y_j\in X \\ -1, & \textnormal{ if } \y_j\in Y \end{cases}$ 
\STATE $\widehat{F}(\z_j) \gets \frac{1}{M}\sum_{\ell=1}^M c_\ell\Phi_n(\z_j,\y_\ell) $ 
\STATE $\rho\gets \textnormal{minimal separation between $\z_j$}$    
\STATE $p\gets 1-\alpha(\rho/n)^q$ 
\STATE $N \gets \log(\alpha) / \log(p)$ 
 \FOR{ $k=1$ to $N$}
\STATE  $\pi \gets Permutation(M)$ 
\STATE $F_k(\z_j) \gets \frac{1}{M}\sum_{\ell=1}^M c_{\pi(\ell)} \Phi_n(\z_j,\y_\ell) $ 
\ENDFOR
\STATE $T(\z_j) \gets Percentile(\{F_k(\z_j)\},p)$ 
\STATE $D(\z_j) \gets \mathlarger{\mathlarger{\mathlarger{\mathbbm{1}}}}  \left(|\widehat{F}(\z_j)| > A\cdot T(\z_j)\right)$ 
\STATE \textbf{Return:}$\widehat{F}(\z_j)$, $D(\z_j)$  
\caption{Algorithm for determining significance of empirical witness function using label permutation.  $\widehat{F}(\z_j)$ is notation for the empirical witness function at $\z_j$, and $D(\z_j)$ is an indicator for whether $\widehat{F}(\z_j)$ is significantly nonzero.}
\end{algorithmic}
\label{permutationalgorithm}
\end{algorithm}

\begin{algorithm}[h]
\begin{algorithmic}[1]
\item[{\rm a)}] \textbf{Input:} Data sets $\{X_i\}_{i=1}^C$, points $Z=\{\z_1, ..., \z_K\}$ at which to inspect significance, level of confidence $A$
\item[{\rm b)}] \textbf{Output:} Estimate of $\widehat{F}(\z_j)$ for all $\z_j\in Z$ and estimate of whether the region is dominated by one class
\STATE $\alpha \gets 0.05$
\STATE $\y \gets \cup_{i=1}^C X_i $ 
\STATE $M \gets \sum_{i=1}^C \left\vert X_i\right\vert$ 
\STATE $c^{(i)}_j \gets \begin{cases}1, & \textnormal{ if } \y_j\in X_i \\ 0, & \textnormal{ otherwise }  \end{cases}$ 
\STATE $\widehat{F}^{(i)}(\z_j) \gets \frac{1}{M}\sum_{\ell=1}^M c^{(i)}_\ell\Phi_n(\z_j,\y_\ell) $ 
\STATE $L_j \gets \arg\max_i \widehat{F}^{(i)}(\z_j) $ 
\STATE $\widehat{F}(\z_j) = \widehat{F}^{(L_j)}(\z_j) - \max_{i\neq L} \widehat{F}^{(i)}(\z_j)$ 
\STATE $\rho\gets \textnormal{minimal separation between $\z_j$}$ 
\STATE $p\gets 1-\alpha(\rho/n)^q$ 
\STATE $N \gets \log(\alpha) / \log(p)$ 
\FOR {$k=1$ to $N$}
\STATE $\pi \gets Permutation(M)$ 
\STATE $\widehat{F}^{(i)}_k(\z_j) \gets \frac{1}{M}\sum_{\ell=1}^M c^{(i)}_{\pi(\ell)} \Phi_n(\z_j,\y_\ell) $ 
\STATE $L \gets \arg\max_i \widehat{F}^{(i)}_k(\z_j) $ 
\STATE $F_k(\z_j) = \widehat{F}^{(L)}_k(\z_j) - \max_{i\neq L} \widehat{F}^{(i)}_k(\z_j)$ 
\ENDFOR
\STATE $T(\z_j) \gets Percentile(\{F_k(\z_j)\},p)$ 
\STATE $D(\z_j) \gets \mathlarger{\mathlarger{\mathlarger{\mathbbm{1}}}}  \left(|\widehat{F}(\z_j)| > A\cdot T(\z_j)\right)$ 
\STATE \textbf{Return:} $\widehat{F}(\z_j)$, $L_j$, $D(\z_j)$
\caption{Algorithm for determining significance of class identifier function using label permutation.  Along with returning $\widehat{F}(\z_j)$ and $D(\z_j)$, it also returns $L_j$, which is the predicted class for $\z_j$. }
\end{algorithmic}
\label{permutationalgorithm_multipleclass}
\end{algorithm}

For all the experimental sections, we will specify several parameters that are necessary to the algorithm.  One parameter is the degree $\mathsf{deg}$ of the polynomials used.  Note that the parameter $n$ in the kernel $\Phi_n$ is given by $n=\sqrt{\mathsf{deg}}$.  We also specify $A$ (the tunable parameter to set the level of confidence for determining significant regions), and the scaling factor on the data $\sigma$.  The scaling factor rescales the data so that $\widetilde{X} = X/\sigma$.  One way to consider this scaling is in analogy with the bandwidth of a Gaussian kernel, where $\exp( - \|\x_i - \x_j\|^2/\sigma^2) = \exp(- \| \widetilde{\x}_i - \widetilde{\x}_j\|^2)$; i.e., the variable $X$ is renormalized to have a variance $1$. In the context of the witness function, $\sigma$ no longer represents a variance parameter, but serves the same role as a free parameter.

\vskip-0.5cm
\bhag{Experiments}\label{exptsect}
\subsection{Toy Examples}\label{toycirclesect}

\begin{figure}
\footnotesize
\begin{center}
\begin{tabular}{ccccc}
\includegraphics[height=.12\textwidth]{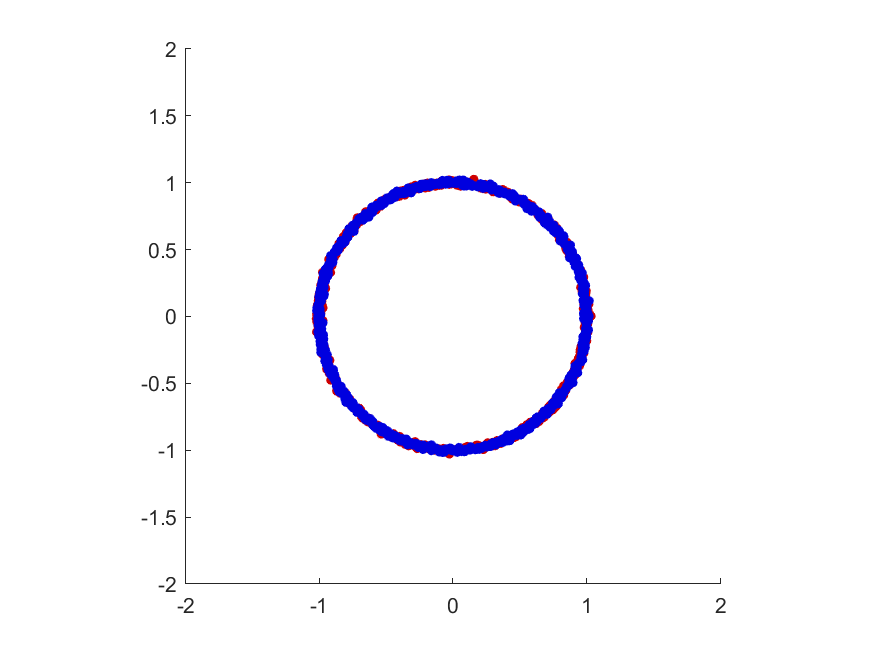} & 
\includegraphics[height=.12\textwidth]{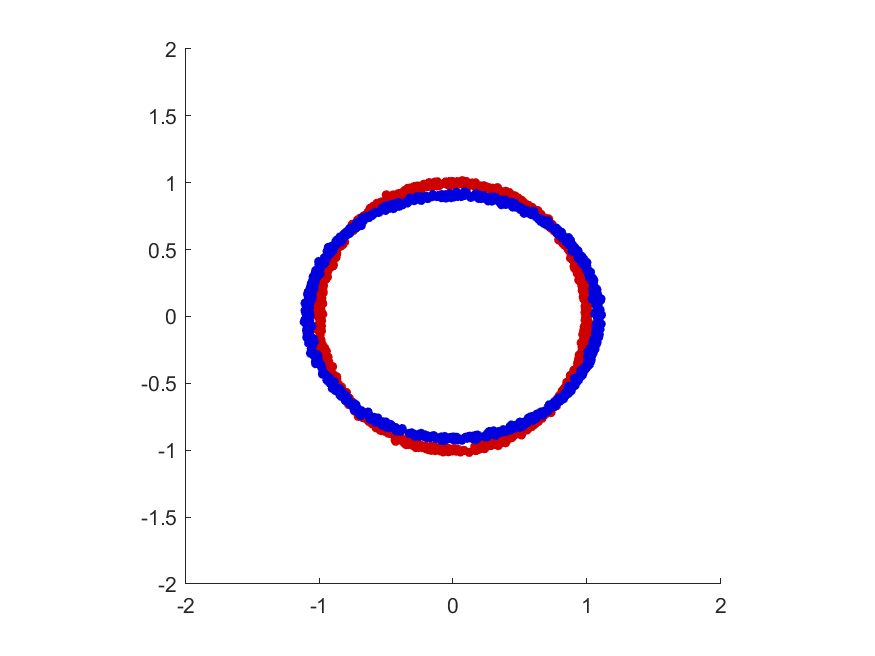} &
\includegraphics[height=.12\textwidth]{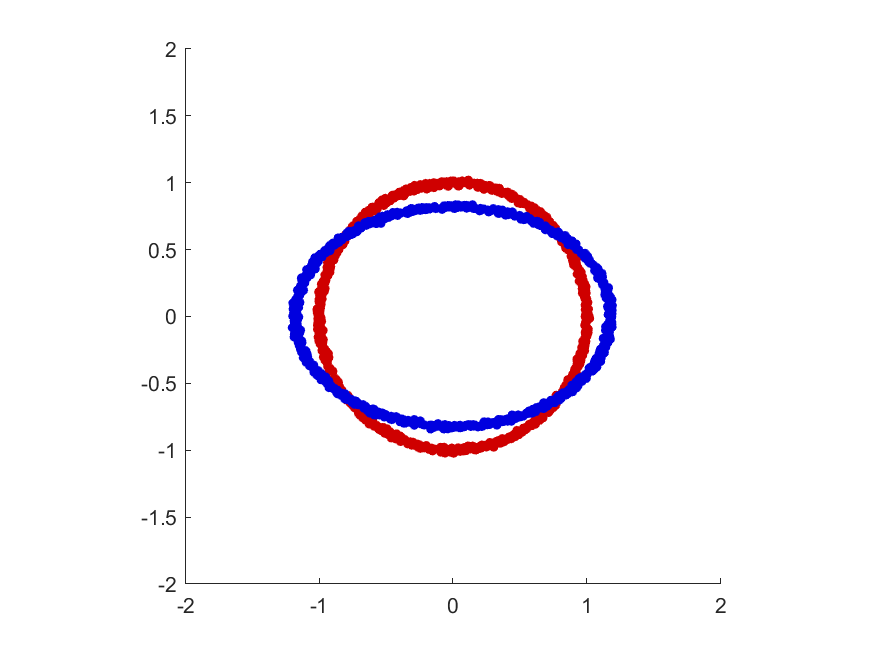} &
\includegraphics[height=.12\textwidth]{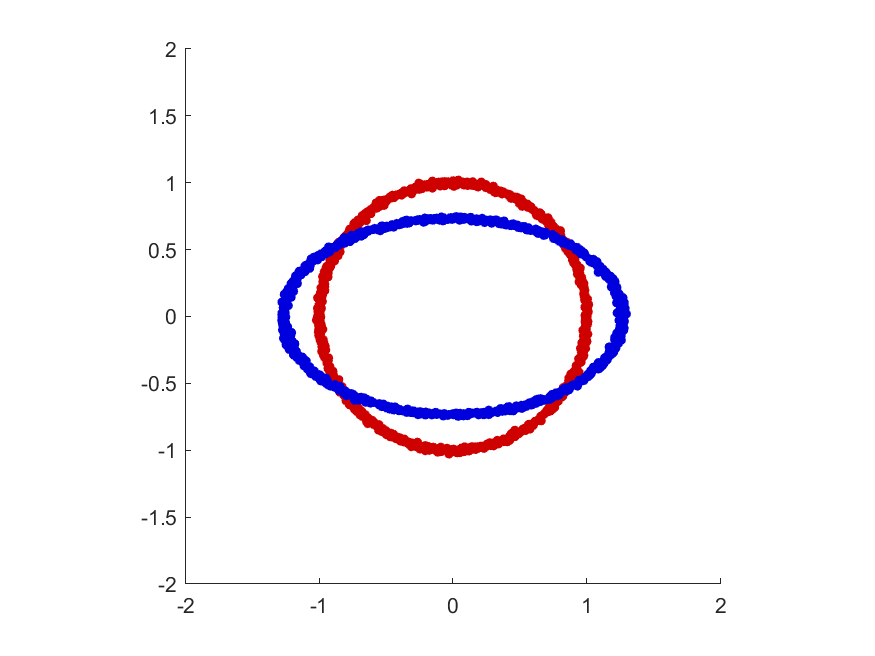} &
\includegraphics[height=.12\textwidth]{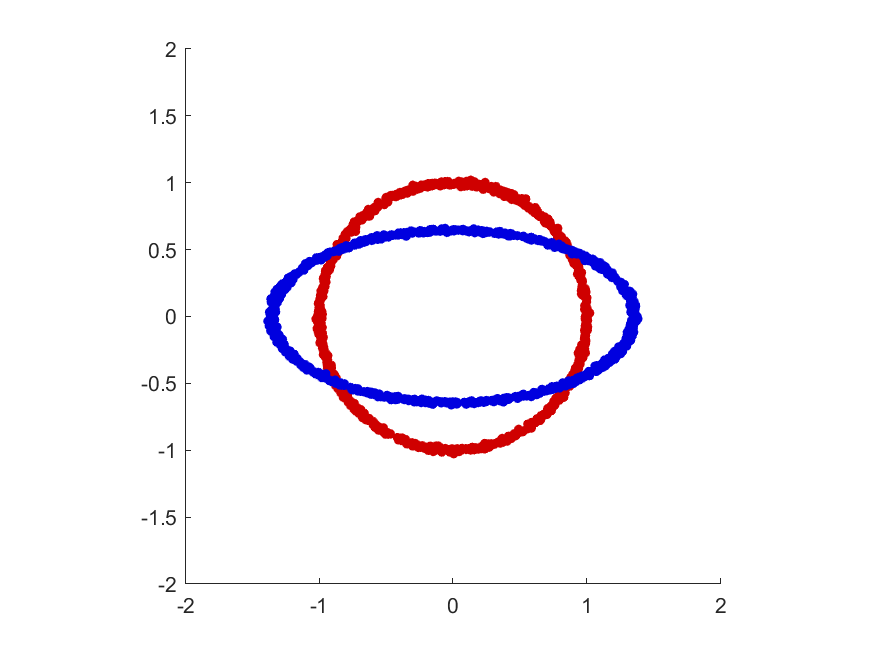} \\
$\varrho=$0 & $\varrho=$0.09 & $\varrho=$0.18 & $\varrho=$0.27 & $\varrho=$0.36\\
\includegraphics[height=.12\textwidth]{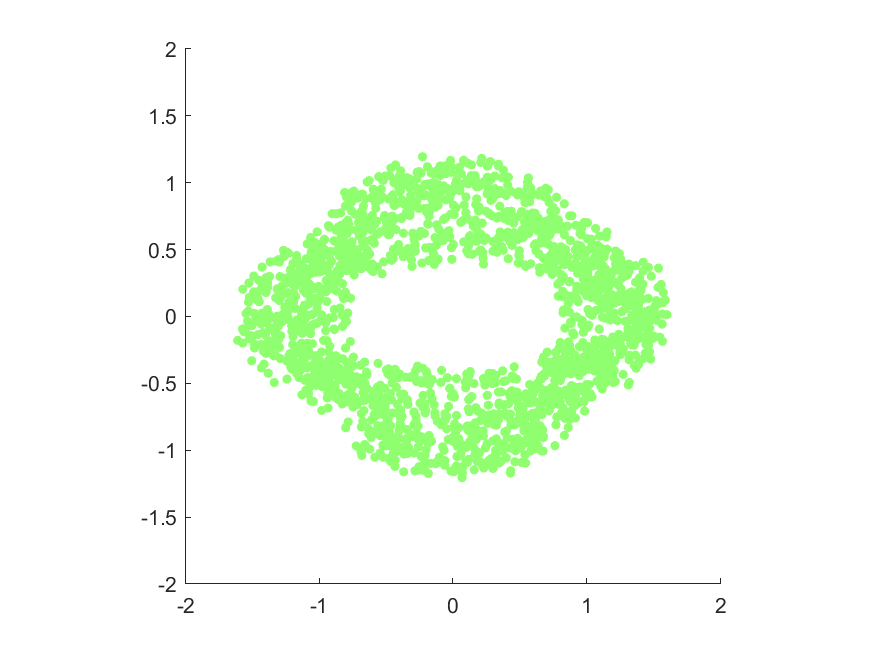} & 
\includegraphics[height=.12\textwidth]{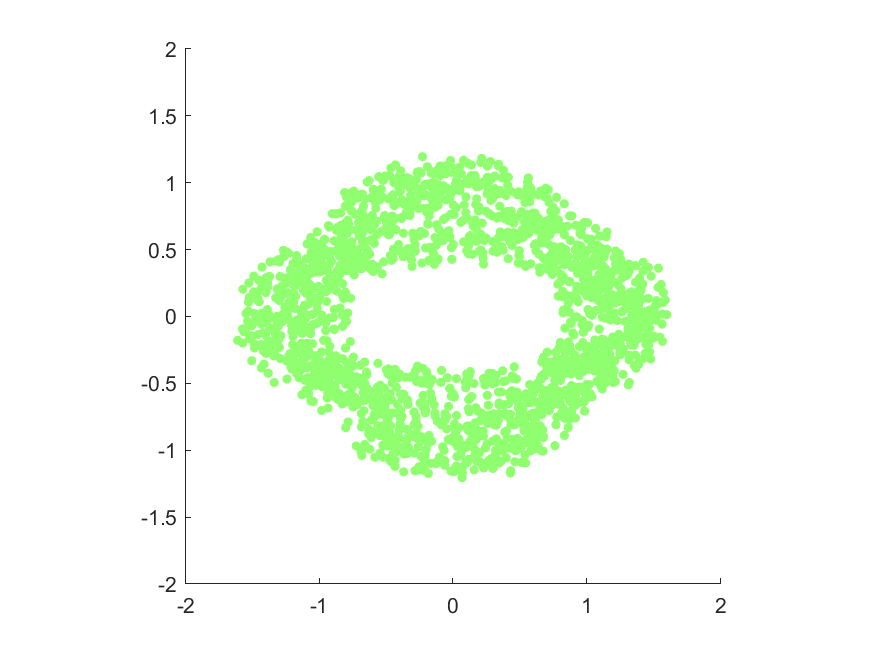} &
\includegraphics[height=.12\textwidth]{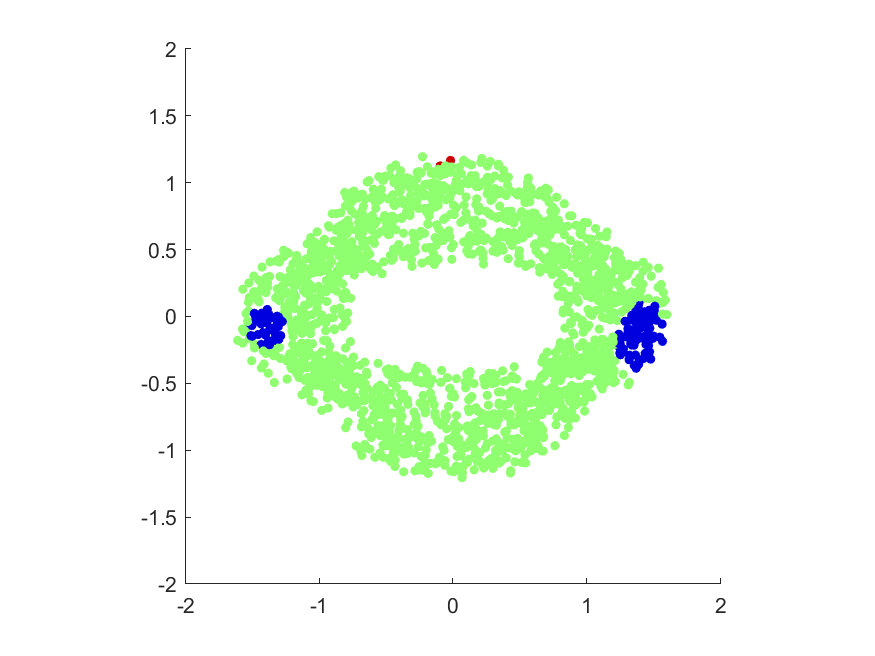} &
\includegraphics[height=.12\textwidth]{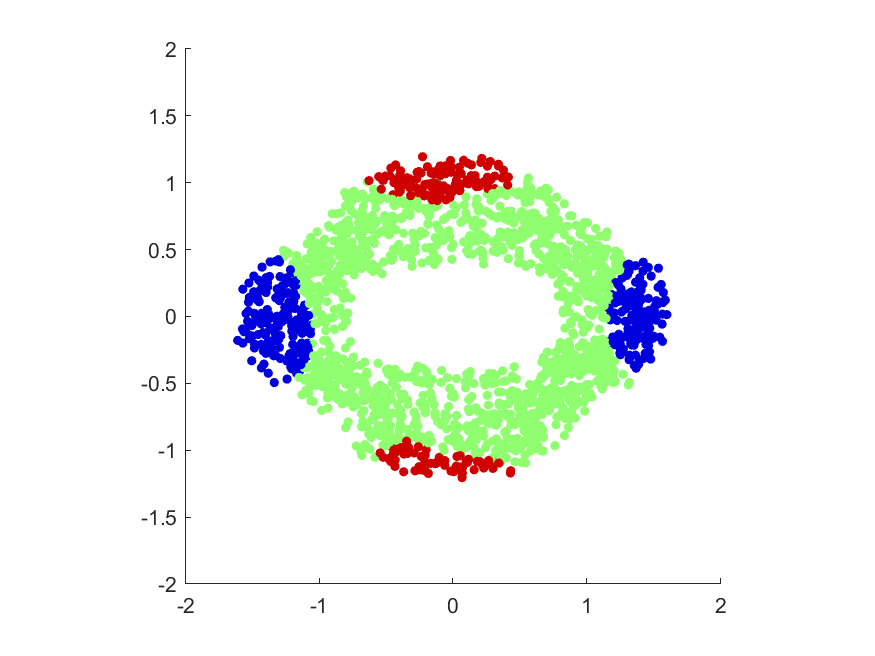} &
\includegraphics[height=.12\textwidth]{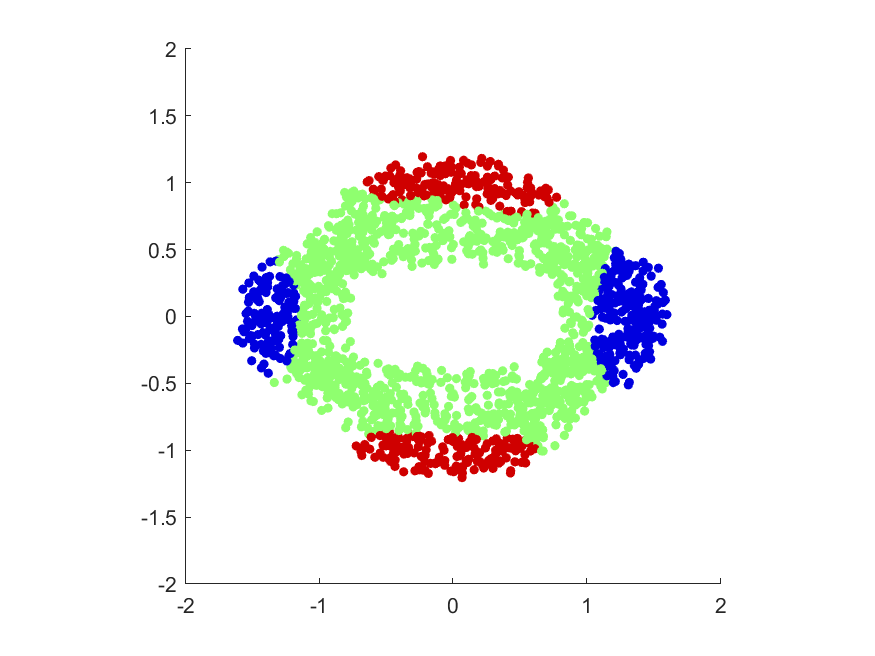} \\
Gauss Sig Wit 0 & Gauss Sig Wit 0.09 & Gauss Sig Wit 0.18 &Gauss Sig Wit 0.27 &Gauss Sig Wit 0.36\\
\includegraphics[height=.12\textwidth]{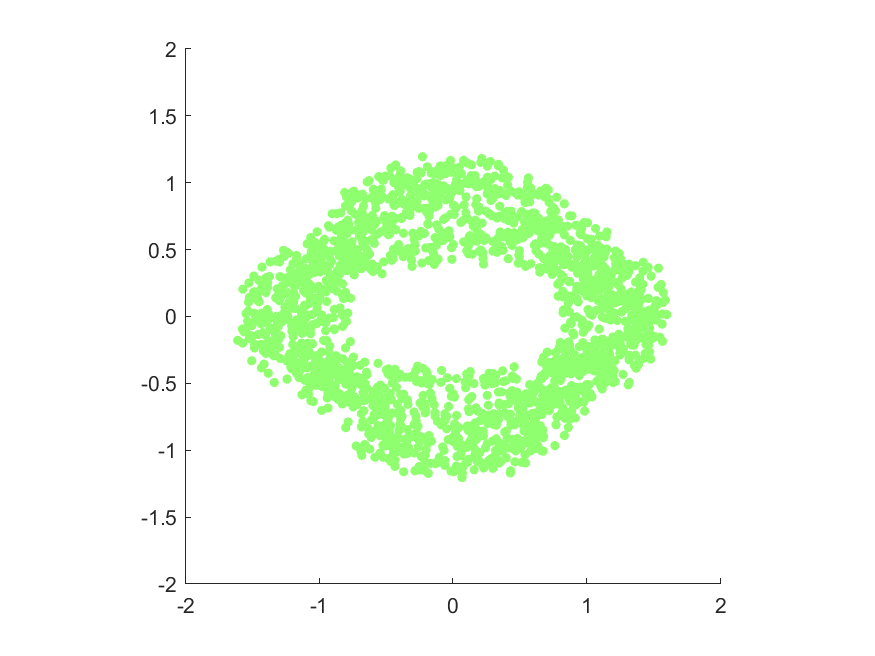} & 
\includegraphics[height=.12\textwidth]{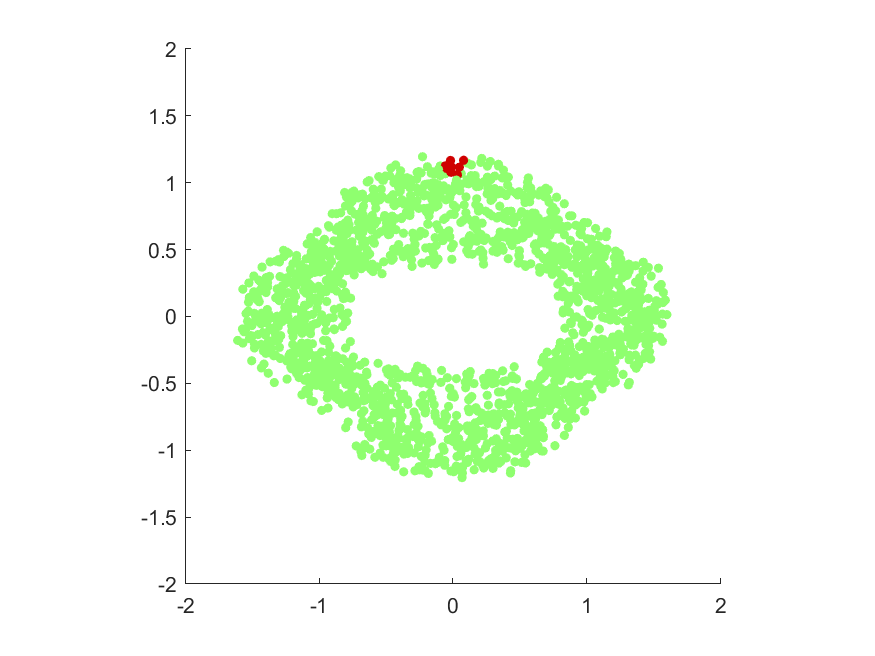} &
\includegraphics[height=.12\textwidth]{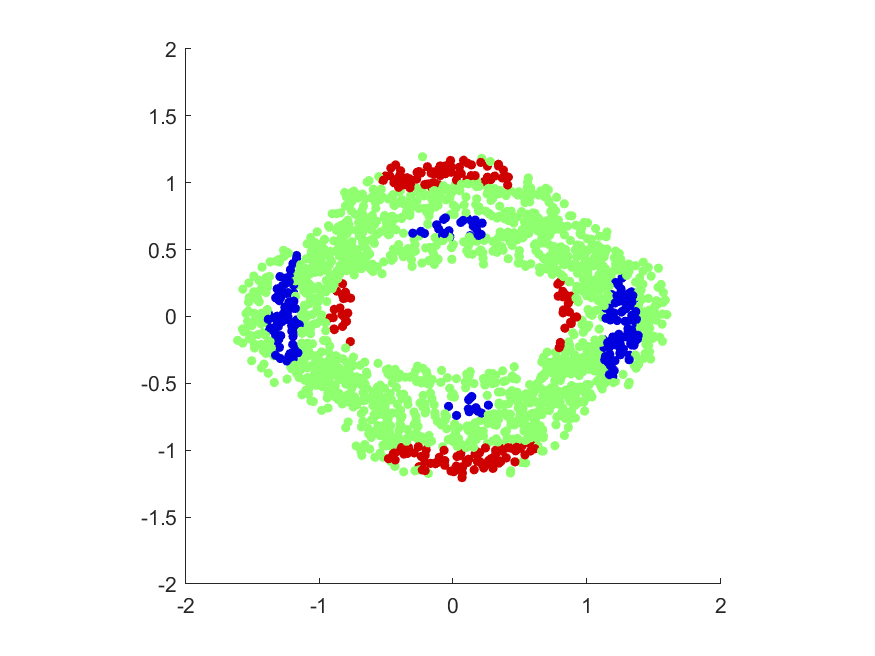} &
\includegraphics[height=.12\textwidth]{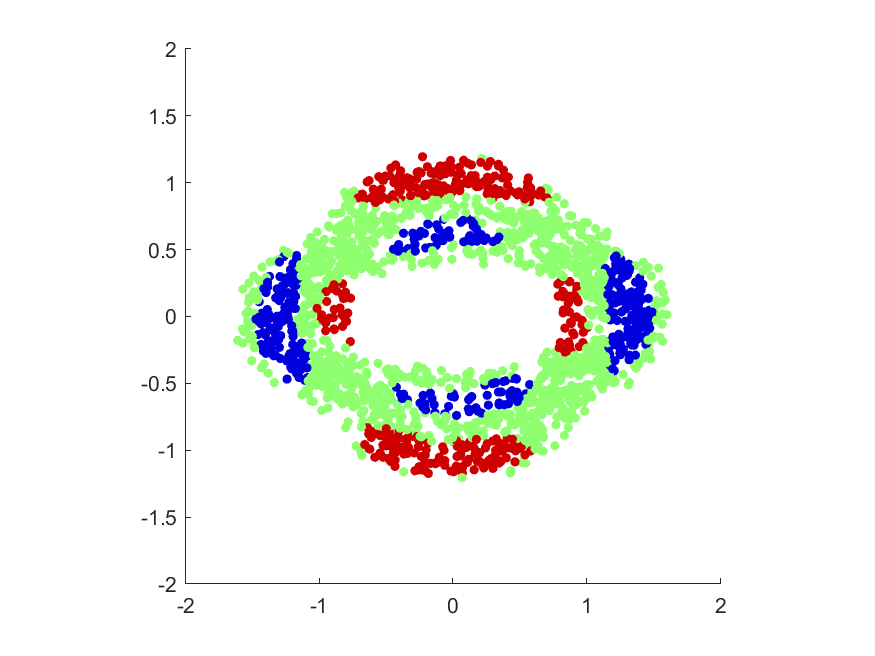} &
\includegraphics[height=.12\textwidth]{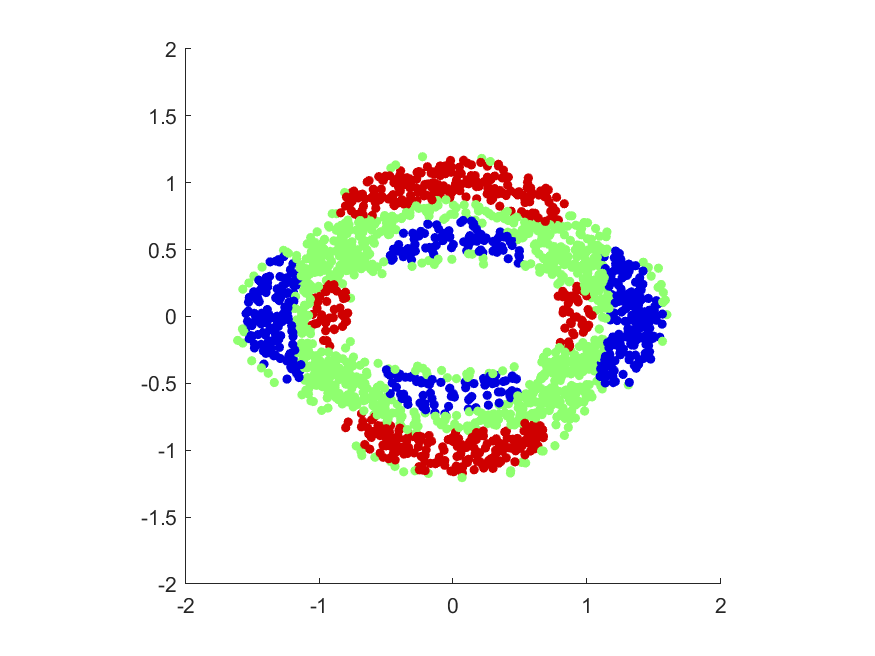} \\
Herm Sig Wit 0 & Herm Sig Wit 0.09 & Herm Sig Wit 0.18 &Herm Sig Wit 0.27 &Herm Sig Wit 0.36\\
\end{tabular}
\caption{(Top) examples of the two data sets with varying $\varrho$, the difference between the principle axis lengths.  (Middle) Points deemed significant by the Gaussian witness function for given $\varrho$.  (Bottom) Points deemed significant by the Hermite witness function for given $\varrho$.  In all figures, green corresponds to $0$, meaning neither distribution dominates.}\label{toycirclefig}
\end{center}
\end{figure}  

We begin the set of experiments with a toy model to demonstrate the benefits of determining the significant regions for the witness function, as well as the benefits of using the Hermite kernel versus using the Gaussian kernel.  
We generate two data sets. 
The first is of the form $(\cos t +\epsilon_1, \sin t+\epsilon_2)$, where $t$ is distributed uniformly on $[0,2\pi)$ and $\epsilon_1,\epsilon_2$ are normal random variables with mean $0$ and standard deviation $0.01$. 
The second data set is of the form $((1+\varrho)\cos t +\epsilon_3, (1-\varrho)\sin t+\epsilon_4)$, where $t$ is distributed uniformly on $[0,2\pi)$ and $\epsilon_3,\epsilon_4$ are normal random variables with mean $0$ and standard deviation $0.01$.
See Figure \ref{toycirclefig} for visualizations of the data sets with various $\varrho$.
We make random draws of $1000$ points of the first form and $1000$ points of the second form for various sizes of $\varrho$, and use Algorithm \ref{permutationalgorithm} to determine the regions of significant deviation.  For this data, we set $\mathsf{deg}=32$, $A=2$, and $\sigma=0.5$.

Figure \ref{toycirclefig} shows the significant areas for varying radii, where the witness function is measured at random points in a ring surrounding the two distributions.  Not only does the Hermite kernel begin to detect significant regions earlier than the Gaussian kernel, the Hermite kernel is also the only kernel to detect the shorter radii of the ellipse.  Also, observe the gap of significance around the points of interaction, in which both distributions are locally similar.

\subsection{Data Generation Through VAE}\label{mnistvaesect}

The next set of experiments revolve around estimating regions of space corresponding to given classes, and sampling new points from that given region.  This problem has been of great interest in recent years with the growth of generative networks, namely various variants of generative adversarial networks (GANs) \cite{goodfellow2014generative} and variational autoencoders (VAEs) \cite{kingma2013auto}.  Each has a low-dimensional latent space in which new points are sampled, and mapped to $\RR^q$ through a neural network.  While GANs have been more popular in literature in recent years, we focus on VAEs in this paper because it is possible to know the locations of training points in the latent space.  A good tutorial on VAEs can be found in \cite{doersch2016tutorial}.

Our first example is the well known MNIST data set \cite{lecun2010mnist}. This is a set of handwritten digits $0\cdots 9$, each scanned as a $28\times 28$ pixel image.  There are $50000$ images in the training data set, and $10000$ in the test data.

\begin{figure}[!h]
\begin{center}
\begin{minipage}{0.2\textwidth}
\includegraphics[height=\textwidth,width=\textwidth]{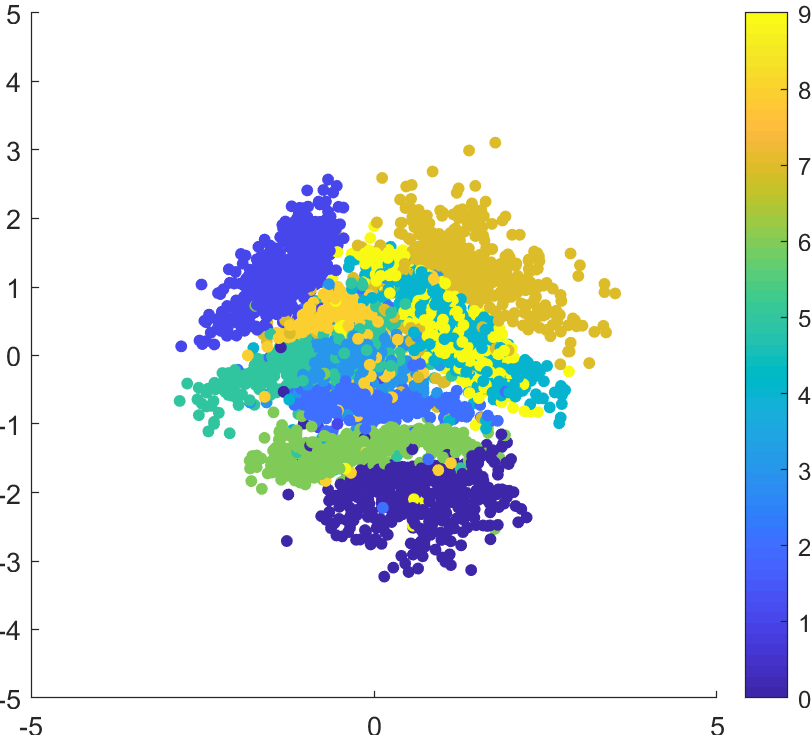}
\end{minipage}
\begin{minipage}{0.2\textwidth}
\includegraphics[height=\textwidth,width=\textwidth]{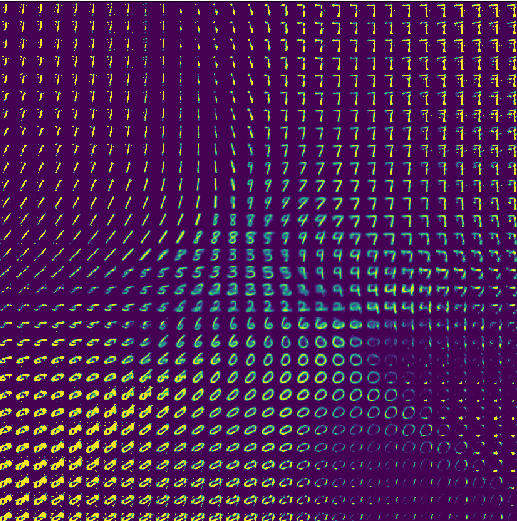}
\end{minipage} 
\begin{minipage}{0.2\textwidth}
\includegraphics[height=\textwidth,width=\textwidth]{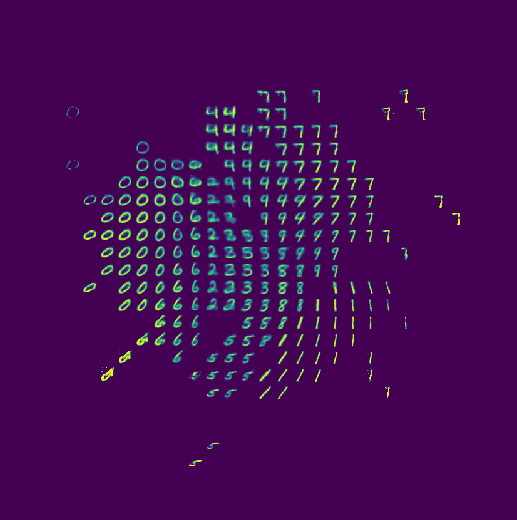}
\end{minipage}
\begin{minipage}{0.2\textwidth}
\includegraphics[height=\textwidth,width=\textwidth]{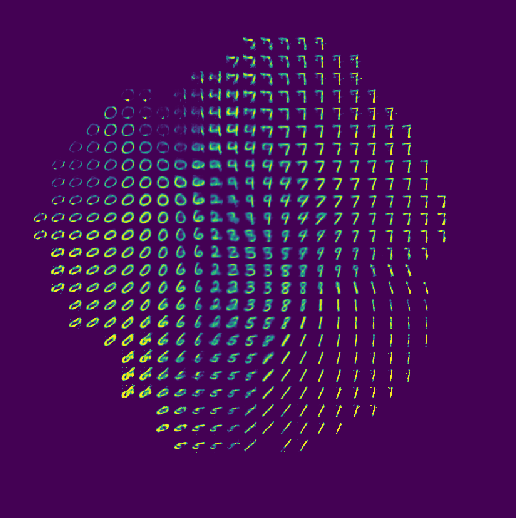}
\end{minipage}
\caption{Left to right: (1) Embedding of training data into 2D VAE latent space.   (2) Reconstructed images from grid sampling in 2D VAE latent space.
(3) Reconstructed images only of grid points that are deemed ``significant'' by the witness function with the Hermite kernel (4) Reconstructed images only of grid points that are deemed ``significant'' by the witness function with the  Gaussian kernel. }\label{fullgridlatentfig}
\end{center}
\end{figure}

In order to select the ``right'' features for this data set, we construct a three layer VAE with encoder $E(\x)$ with architecture $784-500-500-2$  and a decoder/generator $G(\z)$ with architecture $2-500-500-784$, and for clarity consider the latent space to be the 2D middle layer.  In the 2D latent space, we construct a uniform grid on $[-5,5]^2$.  Each of these points can be mapped to the image space via $G(\z)$, but there is no guarantee that the reconstructed image appears ``real'' and no a priori knowledge of which digit will be generated.  We display the resulting images $G(\z)$ in Figure \ref{fullgridlatentfig}, with each digit location corresponding to the location $\z$ in the 2D latent space.  However, we also have additional information about the latent space, namely the positions of each of the training points and their associated classes.  In other words, we know $\z = E(\x)$ for all training data $\x$.  The embedding of the training data $E(\x)$ is displayed in Figure \ref{fullgridlatentfig} as well as the resulting images $G(\z)$ for each $\z$ in the $5\times 5$ grid.
As one can see, certain regions are dominated by a single digit, other regions contain mixtures of multiple digits, and still other regions are completely devoid of training points (meaning one should avoid sampling from those regions entirely).

We use Algorithm~\ref{permutationalgorithm_multipleclass} to determine the ``significant region'' in the embedding space of each class.  In other words, we run Algorithm \ref{permutationalgorithm_multipleclass}  on  $\{E(\x_i)\}_{\x_i \in X}$.  For this data, we set $\mathsf{deg}=128$, $A=2$, and $\sigma=1$.  In Figure~\ref{fullgridlatentfig}, we display the resulting decoded images $D(\x)$ for points $\z_i\in\RR^2$ deemed significant by the Hermite kernel.  Note that most of the clearly fake images from Figure~\ref{fullgridlatentfig} have been removed as non-significant by the witness function.  Figure~\ref{fullgridlatentfig} also computes the same notion of significance with the Gaussian kernel of the same scaling, which clearly has less ability to differentiate boundaries.   We can see in Figure~\ref{fullgridlatentfig} that the Gaussian kernel struggles to differentiate boundaries of classes, and keeps a number of points at the tails of each class distribution.  These points are exactly the points that are poorly reconstructed by the model, as the decoder net hasn't seen a sufficient number of points from the tail regions.

%

\begin{figure}[!h]
\begin{center}
\begin{tabular}{cc}
\includegraphics[height=.2\textwidth,width=.3\textwidth]{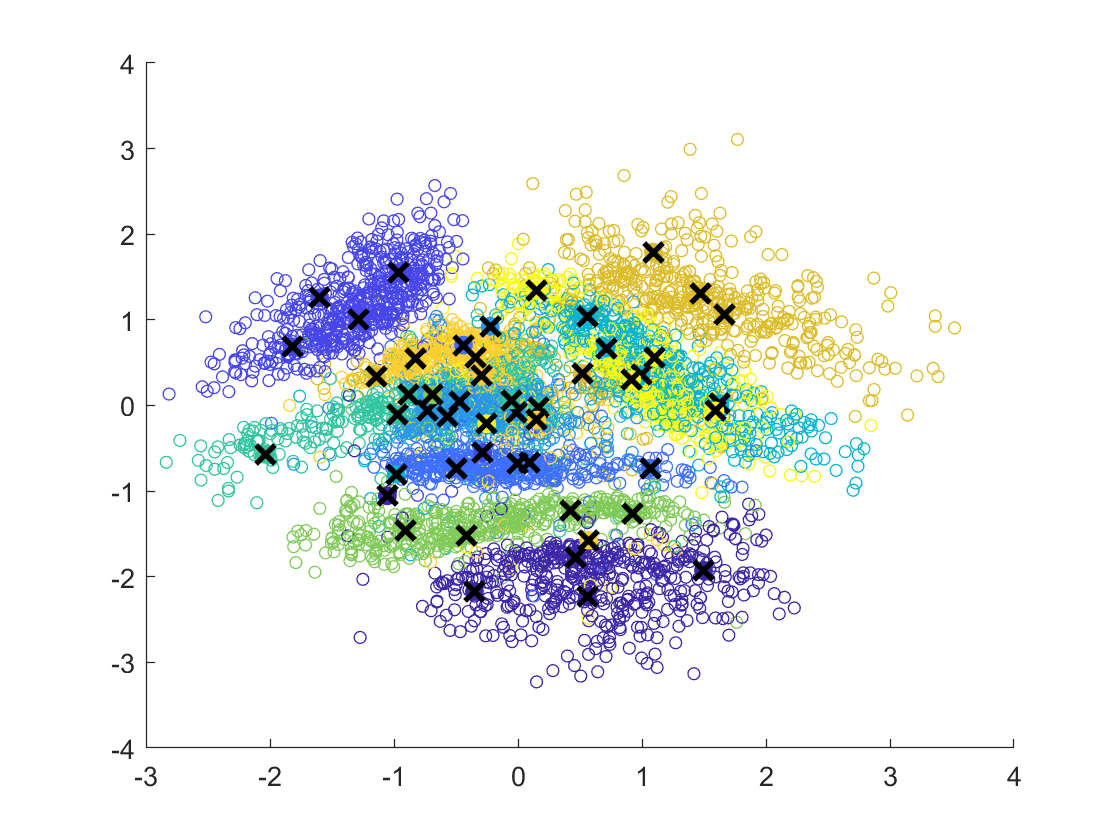} & 
\includegraphics[height=.2\textwidth,width=.3\textwidth]{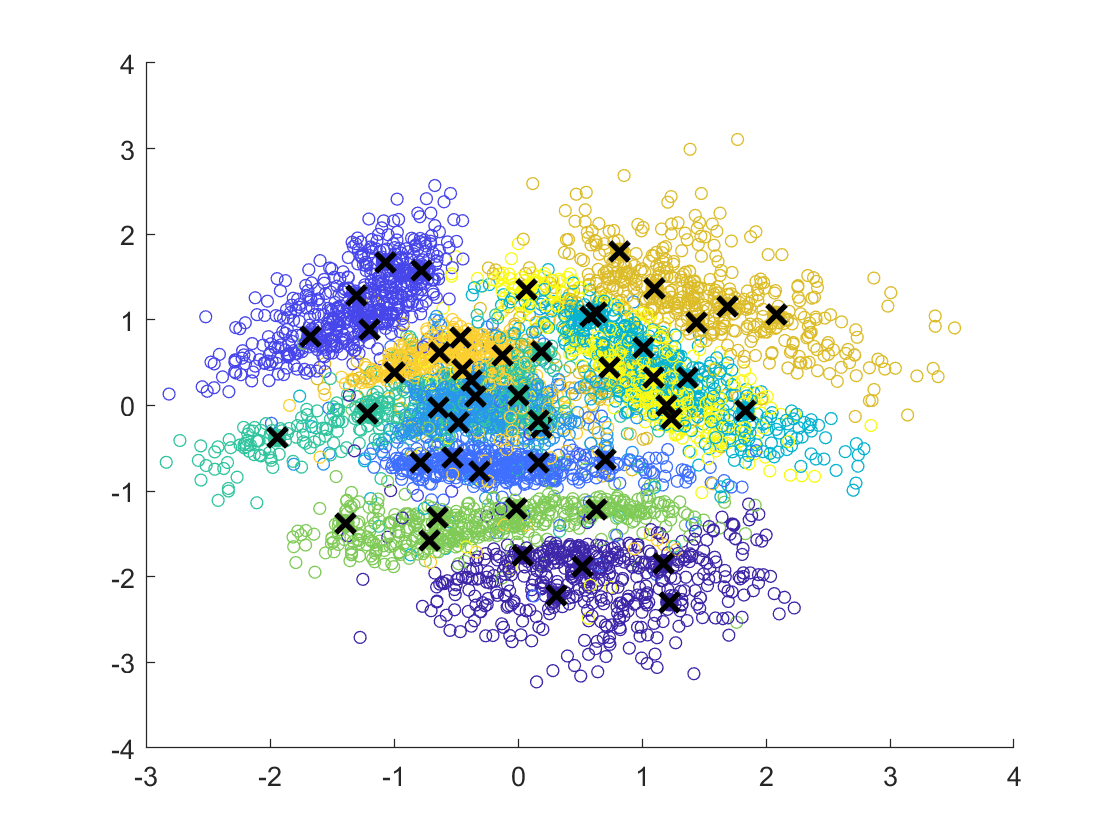}\\
All point GMM centroids & Witness function region GMM centroids \\
\includegraphics[width=.4\textwidth, height=.2\textwidth]{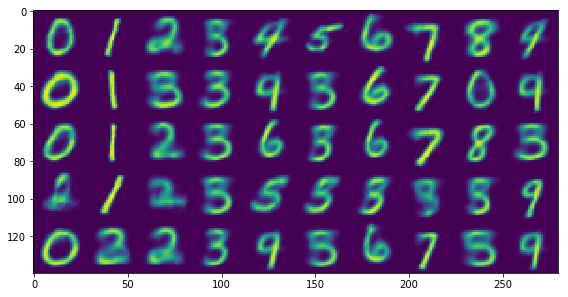} &
\includegraphics[width=.4\textwidth,height=.2\textwidth]{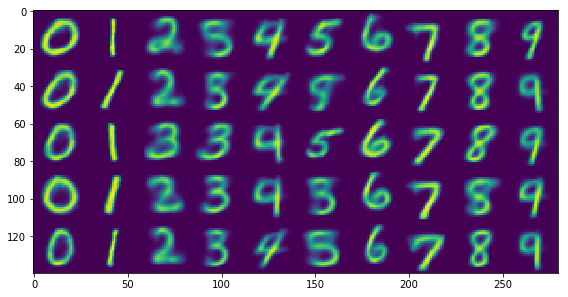}\\
All point GMM centroids reconstructions & Witness function region GMM centroids reconstructions
\end{tabular}
\caption{Prototypical points from each class of MNIST digits, computed from 2D VAE embedding.}\label{mnistwitnessmedianfig}
\end{center}
\end{figure}

We can also use the significance regions to define ``prototypical points'' from a given class.  We do this by fitting the data from each class with a Gaussian mixture model with five clusters.  The means and covariance matrices of each Gaussian are computed through the standard expectation maximization algorithm \cite{reynolds2015gaussian}.
Figure \ref{mnistwitnessmedianfig} shows the centroid values in the embedding of the training data, computed in two different ways.  The first approach is to build the mixture model on all points in a given class.  In other words, we build a Gaussian mixture model with five clusters on the data $\{E(\x_i) : z_i = j\}$ for each of $j\in\{0,1,...,9\}$.  
The second approach is to build the mixture model for each class using only those points that are deemed significant by the witness function test.  
In other words, a mixture model on the restricted dataset $\{E(\x_i) : z_i = j \textnormal{ and } D(\x_i) = 1\}$ for each of $j\in\{0,1,...,9\}$.  Due to the structure of the two-dimensional embedding, some of the mixtures for entire classes are pulled toward the origin by a few outliers from the class that are spread across the entire space.  This causes overlap between the centroids of the classes considered more difficult to separate in a 2D embedding (4's, 6,'s, 9's), and causes problems in the reconstruction.  The centroids of the mixtures for significant regions, on the other hand, have a tendency to remain squarely within neighborhoods of the same class, and their reconstructions $G(\z_i)$ are much clearer.

\subsection{Determining Signficant Class Regions for CNNs}\label{CNNsect}

In this section, we consider learning regions of uncertainty in hidden layers of a convolutional neural network.  Assessing the uncertainty of classification is an important aspect of machine learning, as certain testing points that are far away from training data, or on the boundary of several classes, should not receive a classification.  Even at the last hidden layer of the neural network, there exist points that fall into uncertain regions or boundaries between regions.  Our goal is to examine this last hidden layer and prospectively remove uncertain points.  In doing so, the goal would be to reduce the set of testing points \emph{without a priori knowledge} of ground truth on the testing data in a way that reduces to final classification error on the test set.

We use the last layer of the VGG-16 pretrained CNN that has been rescaled to the CIFAR10 data set, where the last layer contains $q=512$ dimensions.  The CIFAR10 data set is a collection of 60,000 32x32 color images in 10 classes (airplane, bird, cat, deer, etc.), with 6000 images per class.  There are 50000 training images and 10000 test images \cite{krizhevsky2014cifar}. VGG-16 is a well known neural network trained on a large set of images called Imagenet, which is rescaled to apply to CIFAR10.  VGG-16 has  12 hidden layers, and the architecture and trained weights can be easily downloaded and used \cite{simonyan2014very}.   VGG-16 attains a prediction error of $\sim 6\%$ on the testing data.   Our goal is to detect the fraction of images that are going to be misclassified prior to getting their classification, and thus reduce the prediction error on the remaining images.

\begin{figure}
\centering
\begin{tabular}{ccc}
\includegraphics[width=.3\textwidth]{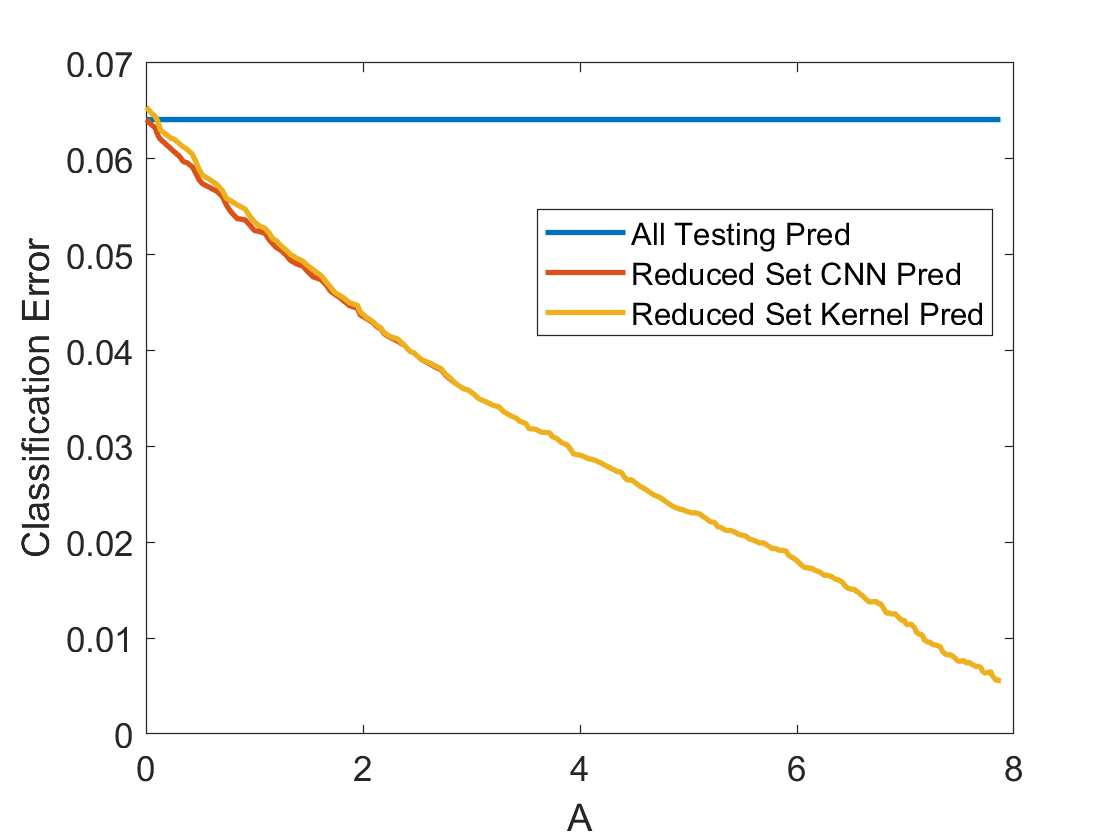} & 
\includegraphics[width=.3\textwidth]{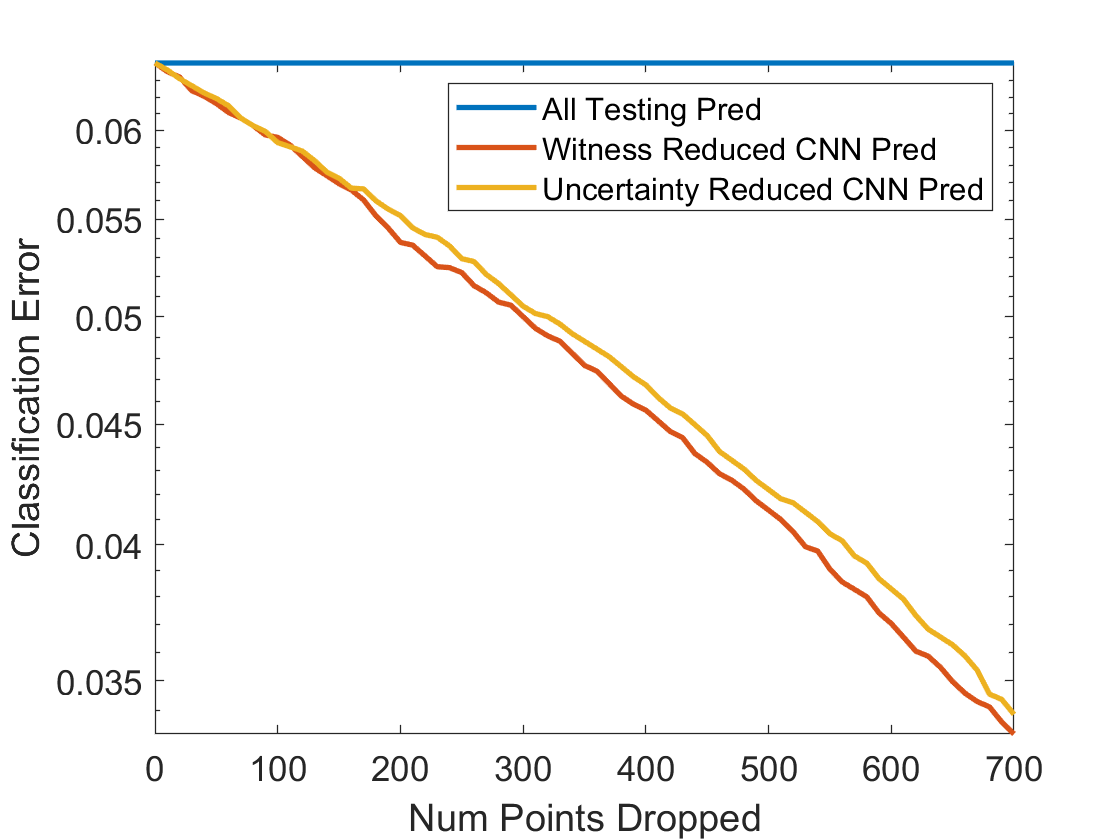} & 
\includegraphics[width=.3\textwidth]{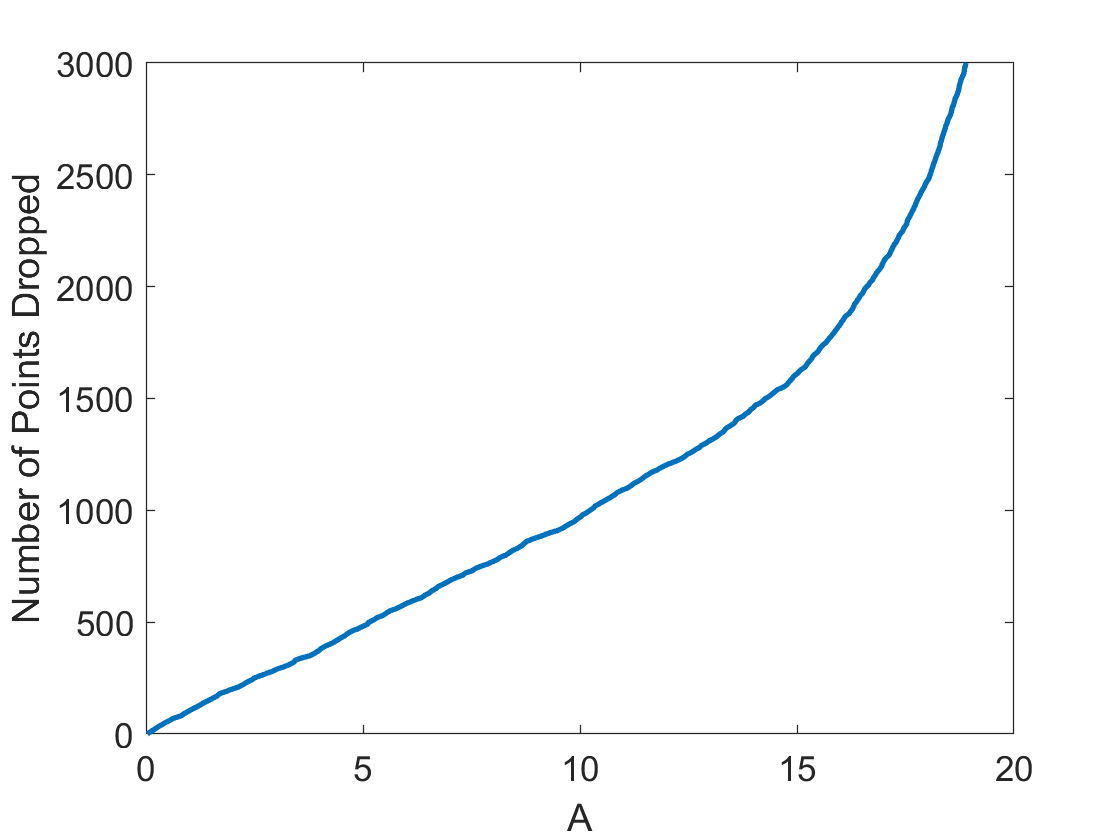} 
\end{tabular}
\caption{(Left) Classification error on points deemed ``significant'' as a function of $A$.  (Middle) Classification error as a function of the number of points removed for being ``uncertain'', for both the witness function and the size of gap of CNN outputs.  These two measures are placed on the same scale by considering the number of points removed.  (Right) The relationship between the number of points dropped and parameter $A$ in our algorithm.}\label{cifarwitnessfig}
\end{figure}

  We create the  witness function on the testing data embedded into this final hidden layer, and determine the threshold by permuting the known labels of the training data.  For this data, we set $\mathsf{deg}=128$ and $\sigma=7$ ($\sigma$ was chosen as the median distance to the $100^{th}$ nearest neighbor in the training data).  The parameter $A$ is not set in this section as we are varying it across the data.
  Figure \ref{cifarwitnessfig} shows the decay of the classification error as a function of $A$, which has a direct correspondence to the overall probability of error.  While there is a reduction of the overall number of testing points, the set of points that remain have a smaller classification error than the overall test set.  We also show on this reduced set that the label attained by the final layer of the CNN and the estimated label attained by taking the maximum estimated measure across all $10$ classes are virtually identical.

Figure \ref{cifarwitnessfig} also compares the decay of the classification error to the uncertainty in classification as defined by the last layer of the CNN.  A CNN outputs a vector, which we call $g(\x)$, of the probability a point lies in each of the classes.  A notion of uncertainty in the network can be points that have the smallest gap between the prediction of the most likely class  $L = \arg\max_{i} g_i(\x)$ and the second highest classification score, which yields an certainty score $g_L(\x) - \max_{j\neq L} g_j(\x)$.  We sort the testing pionts by this gap, and remove the first $k$ points with the smallest gap.  
The larger this gap is, the more certain the CNN should be about the correct classification.  As shown in Figure \ref{cifarwitnessfig}, our witness function method yields a quicker decay in the classification error as a function of the number of points removed. 

A benefit of our approach to quantifying uncertainty is that it explicitly demonstrates that the points classified poorly are those that sit at the boundary between class clusters in the last hidden layer of the network.  This means that even at the last layer of the network, misclassified points are still considered ``outliers'' by the class distributions to which they lie closest.  

\subsection{Term Document Significance}\label{SNembeddingsect}
In this section, we consider term-document organization and characterizing types of words that are indicitive of a class of documents.  We use the Science News dataset, which consists of 1046 articles across 8 categories, and their use of 1153 popular words that appear in the magazine.  The categories are: Anthropology, Space, Behavioral Psych, Environmental Science, Life Science, Math/CS, Medicine, and Physics/Tech.  There are obvoius overlaps of these categories, so not every article in a category will necessarily contain ``indicative words'' of that category alone.  

\begin{figure}[!h]
\begin{center}
\begin{tabular}{cc}
\includegraphics[height=.3\textwidth]{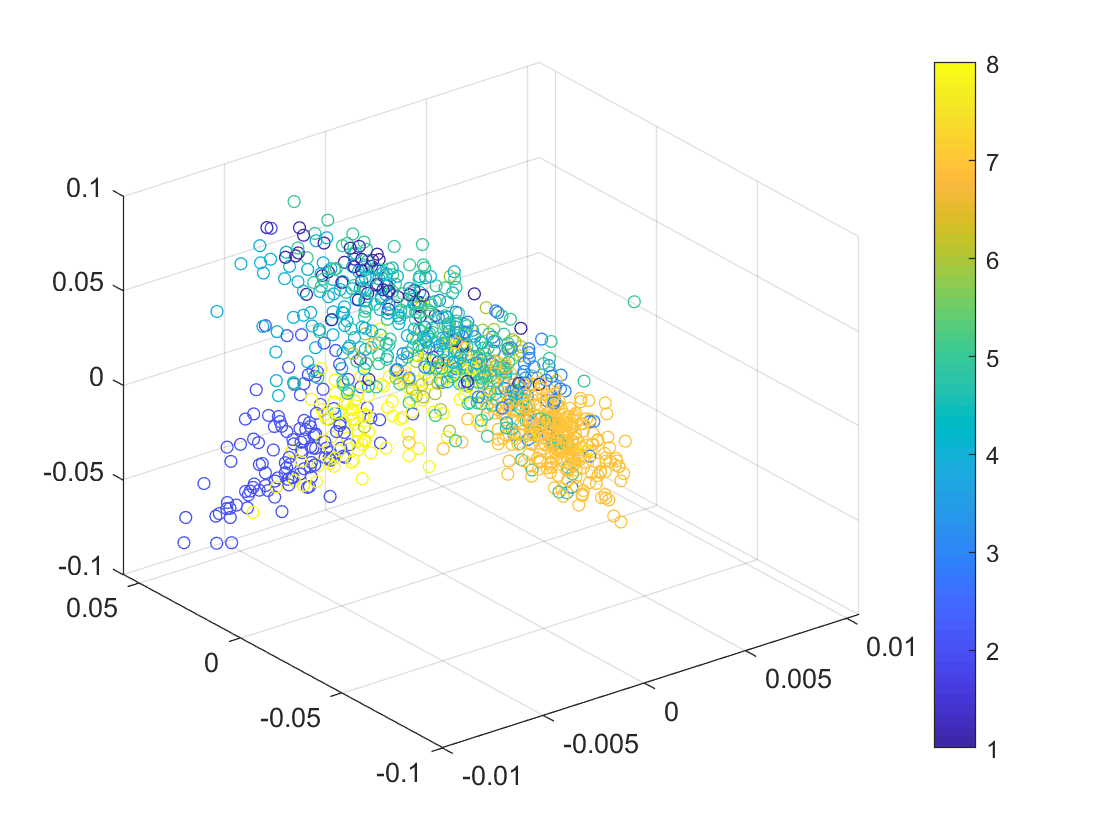} & 
\includegraphics[height=.3\textwidth]{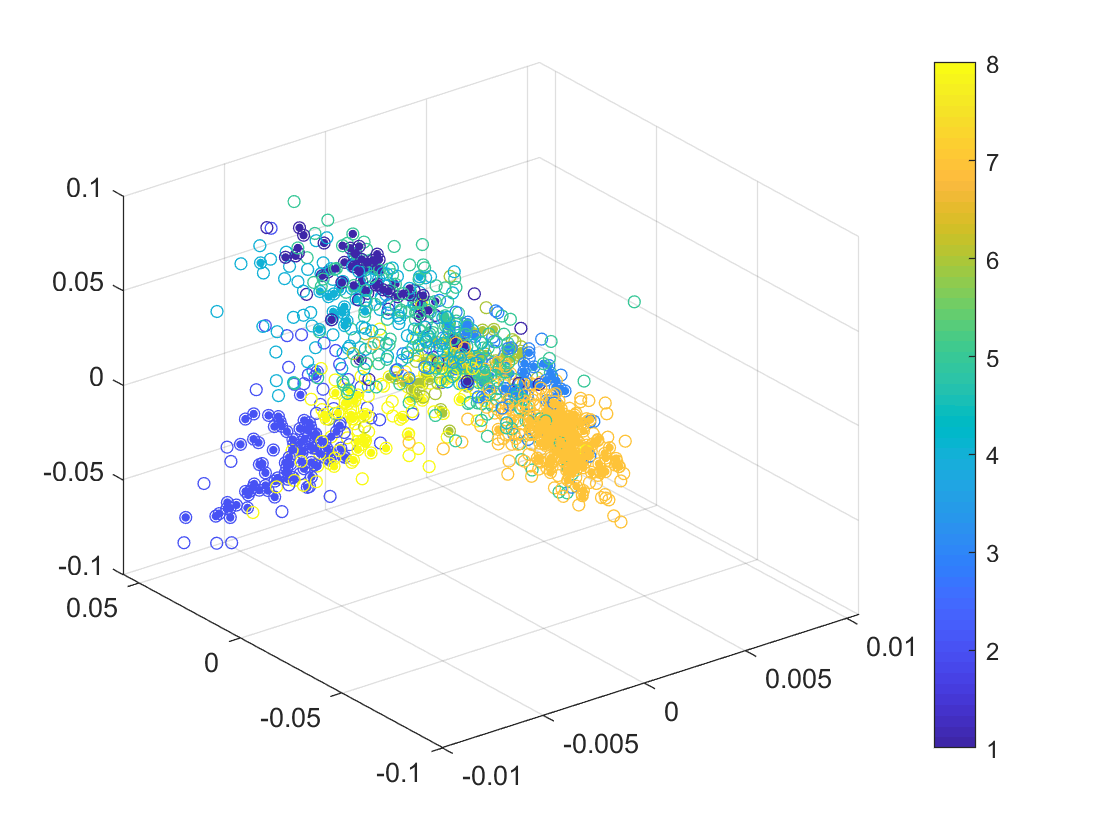}
\end{tabular}
\caption{(Left) Hierarchical topic embedding of documents.  (Right) Embedding highlighted by grid points deemed significantly within one class.}\label{SNembeddingfig}
\end{center}
\end{figure}

We begin by taking the top 3 principle components of the words, and generating a hierarchical topic modeling by splitting the words into four levels of 4, 8, 16, and 32 clusters respectively.  Each document at a given clustering level is encoded by the fraction of its words that fall into a given cluster, and the new features of a document become these histograms across all 4 levels of word splits.  From here, we take the top three principle components of the documents to create a low-dimensional embedding of all documents.  This embedding is displayed in Figure \ref{SNembeddingfig}.   We then run Algorithm \ref{permutationalgorithm_multipleclass} to determine the significant regions for each class.  For this data, we set $\mathsf{deg}=32$, $\sigma=1$, and $A = 2$.

We sample the embedding at 10000 random grid points in the embedding space, and display the significant region in Figure \ref{SNembeddingfig}.  It is important to note the meaning of these regions of significance, namely that rejection of the null hypothesis at a given point indicates that the concentration of points from one class in that region is well beyond any concentration that would occur due to chance.  

%

In an effort to quantify the significant regions and their benefits, we designed the following simple experiment.  For every point, we compare its class to the class of its nearest neighbor, and we record the average classification score across all classes.  Namely, for data $X$ and corresponding labels $Y$, we compute
\begin{eqnarray*}
\EE_{\x_i\in X}[ \delta(y_i, y_{\{\textnormal{NN of $\x_i$ in $X$}\}})], & \textnormal{ and } & \EE_{\x_i\in X} [\delta(y_i, y_{\{\textnormal{NN of $\x_i$ in centroids from $X$}\}})],
\end{eqnarray*}
where $\delta(\cdot,\cdot)$ is a dirac delta function.  We run this experiment for $X$ being all documents, and for $X$ being ``significant'' documents as deemed by the witness function.  The results are in Table \ref{termdocneighbortable}.  Clearly, restricting ourselves to the documents deemed significantly within one class greatly increases the reliability of the neighborhood and the computed centroids of the classes.
\begin{table}[!h]
\begin{center}
\begin{tabular}{|c|c|c|}
\hline
 & Neighbor in given set of docs & Centroid computed from given set of docs \\
 \hline
All documents &   0.5143 &   0.5344      \\
Sig. documents &  {\bf 0.7156 } & {\bf  0.7635  } \\
\hline
\end{tabular}
\end{center}
\caption{Nearest neighbor classification of Science News documents across all documents and across only significant documents.}\label{termdocneighbortable}
\end{table}

\subsection{Propensity Matching and Non-Experiment Sampling}\label{lalondesect}
As a final example, we consider the problem of propensity matching and scoring.  In this setting, we consider two sets of observable (e.g. non-randomized) data in which one set was given a treatment and the other was not (which serves as a control set).  There exists questions around how to determine exactly where these two datasets disagree with one another, and which data points to remove because they are biasing either the treated or control groups (i.e. with their features, they were virtually guaranteed to be either in treatment or in control, and we can't extrapolate treatment effectiveness for them).    This is traditionally done with different versions of logistic regression and matching observations with approximately equivalent probabilities of treatment \cite{rubin1996matching}.

We address this problem in the context of the  canonical LaLonde data set \cite{lalonde1986evaluating}.  This is a data set of men in the National Supported Work Demonstration who were either given (or not given) on job training for $>9$ months.  The ultimate goals of this data are to determine the monetary benefits of job training, but we will focus on detecting differences between the groups that were and were not treated.  The pre-treatment features of the people are age, education, Black (1 if black, 0 otherwise), Hispanic (1 if Hispanic, 0 otherwise), married (1 if married, 0 otherwise), nodegree (1 if no degree, 0 otherwise), RE74 (earnings in 1974), and RE75 (earnings in 1975).  We choose a subset of the data that has information on RE74 following the work of Dehejia-Wahba \cite{dehejia1999causal}.   This leaves us with $260$ control observations and $185$ treated observations.

After z-scoring the $8$ dimensional data (i.e. subtract mean and divide by standard deviation), we apply Algorithm~\ref{permutationalgorithm}    comparing the data to itself (rather than constructing a grid in 8D space).  We use $\mathsf{deg}=16$, $\sigma=1$, and $A=2$. Here we take $\widehat{F}>0$ to be treated and $\widehat{F}<0$ to be control.  
Table \ref{lalondetab} shows the means of $SigTreat = \{\x_j: \widehat{F}(\x_j)>0 \textnormal{ and } D(\x_j)=1\}$ and $SigControl = \{\x_j: \widehat{F}(\x_j)<0 \textnormal{ and } D(\x_j)=1\}$, as well as the means of the treated and control groups independent of significance of the witness function.

\begin{table}[!h]
\centering
\begin{tabular}{|l|cccccccc|}
\hline
 & Age & Years Ed. & Black & Hispanic & Married & No Degree & RE74 & RE75\\
\hline
Control Means & 25.1 & 10.1 & 0.83 & 0.11 & 0.15 & 0.83 & 2107.0 & 1266.9 \\
$SigControl$ Means & 20.7 & 10.4 & 0.33 & 0.67 & 0.00 & 0.89 & 6325.9 & 1658.2 \\
Treated Means & 25.8 & 10.3 & 0.84 & 0.06 & 0.19 & 0.71 & 2095.6 & 1532.1 \\
$sigTreat$ Means  & 30.6 & 12.4 & 1.00 & 0.00 & 0.00 & 0.00 & 9351.8 & 2525.9\\
\hline
\end{tabular}
\caption{Mean value of each feature for the control group and treated group as a whole, and for the subsets of the groups that are deemed to be definitively within one class.}\label{lalondetab}
\end{table}

Table \ref{lalondetab} tells a clear story, namely that the people unlikely to be given job training (i.e. people in $SigControl$) are 
young, average educated, Hispanic men that are not married, likely don't have a degree, and dropped considerably in income between 1974 and 1975.  
On the other hand, people that were almost guaranteed to be given job training (i.e. people in $SigTreat$) are
older, above average educated, black men that are unmarried, have a degree, and previously made well above average income.  
These results are consistent with the biases being offered job training that are identified in previous research on propensity matching \cite{dehejia1999causal}.

\begin{figure}
\centering
\begin{tabular}{cc}
\includegraphics[width=.45\textwidth,height=.2\textwidth]{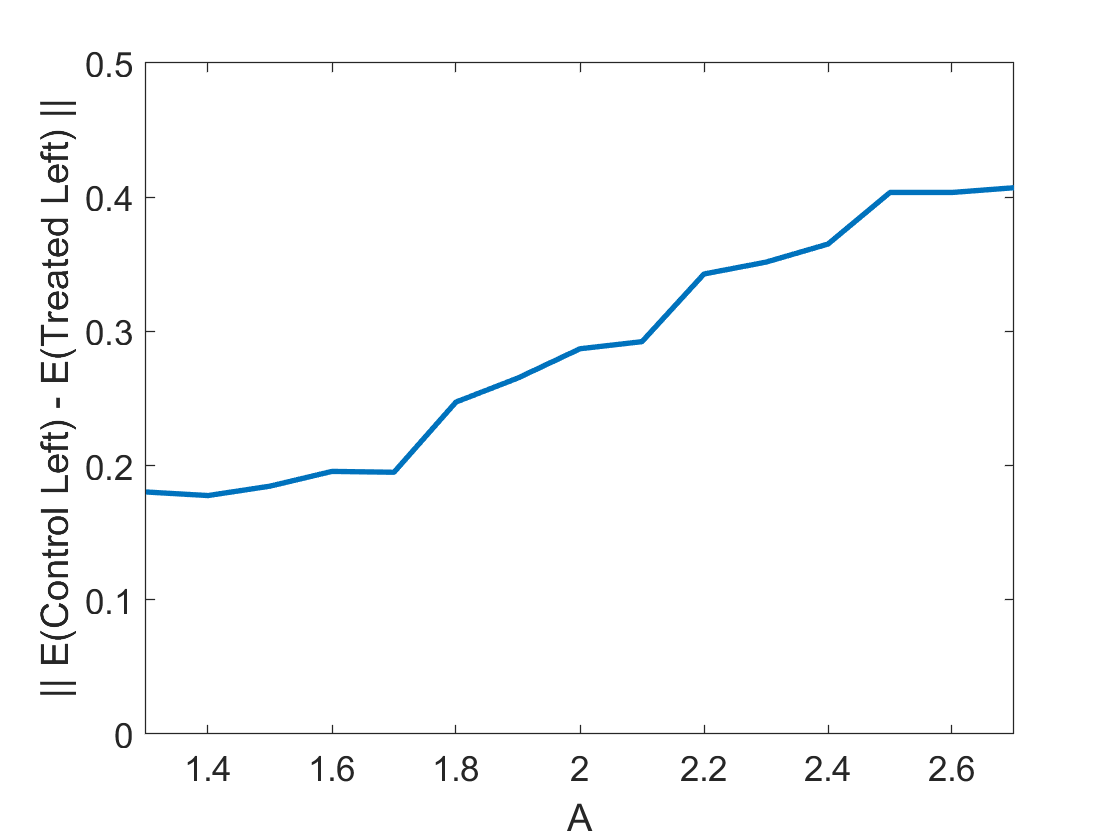} & 
\includegraphics[width=.45\textwidth,height=.2\textwidth]{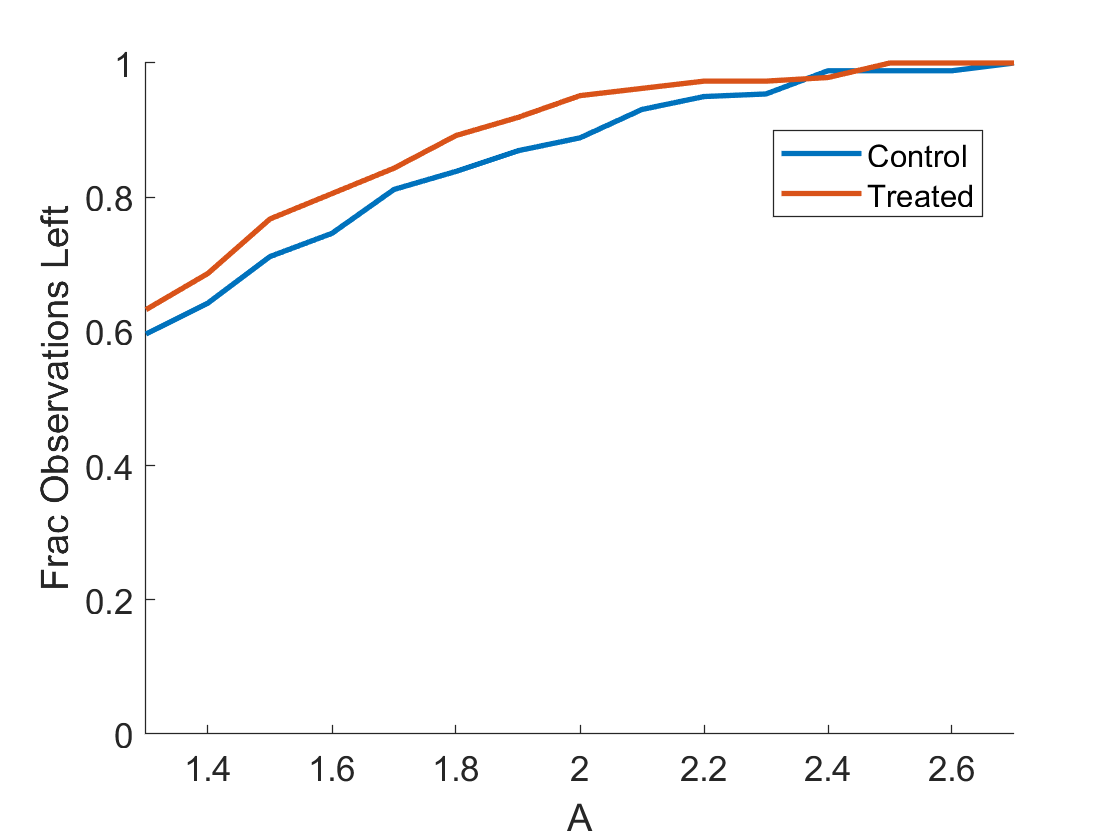}
\end{tabular}
\caption{(Left) Difference of means of control and treated groups for LaLonde data after removing points with $D(Z)=1$ for varying $A$.  (Right) Fraction of observations left in control and treated groups for LaLonde data after removing points with $D(Z)=1$ for varying $A$.   For both plots, $A$ closer to $1$ imples we are liberal with the points being removed, meaning the sets of points remaining will be small but much more similar between treated and control.}\label{lalondefig}
\end{figure}

We also display in Figure \ref{lalondefig} the balancing that occurs after removing those people that fall significantly into one group or the other, and display this as a function of $A$ in Algorithm \ref{permutationalgorithm}.  We compare the mean of the remaining control group and the mean of the remaining treated group, and plot the $\ell^2$ norm difference between these mean vectors.  
This demonstrates that, as we remove observations deemed to be significantly in one class or the other, the remaining groups move closer together in mean and become more balanced.  

\bhag{Proofs}\label{pfsect}
The basic idea behind the proof of Theorem~\ref{locprobesttheo} is the following. 
Theorem~\ref{localchartheo} gives a deterministic estimate on $\|\sigma_n(Ff)-Ff\|_{\mathbb{B}(\x_0,r)}$. 
In turn, \eref{iterexp} expresses $\sigma_n(Ff)$ as an expected value expression, and $\widehat{F}$ is clearly an estimator of this expression. 
Therefore, at each point $\x$, one can estimate $\widehat{F}(\x)-\sigma_n(Ff)(\x)$ using Hoeffding's inequality. 
We note that this difference is a weighted polynomial of degree $< n^2$. 
To convert this estimate into a norm estimate, we need to 
obtain a finite subset of the ball around $\x_0$ so that the norm of any weighted polynomial of degree $<n^2$ is of the same order of magnitude as the norm on the points of this finite subset, which is therefore called a norming subset.
While Corollary~\ref{coveringcor} gives an insight in this direction, a main technical difficulty here is that the norm of the gradient of a weighted polynomial on a cube needs to be estimated by the norm of the polynomial on the cube itself.
If one allows the bounds to depend upon the point $\x_0$, then this is a trivial consequence of the Markov inequality. 
A much deeper argument is needed to obtain the bound independent of $\x_0$. 
An inequality of this sort is known as the Videnskii inequality proved in the context of trigonometric polynomials on arcs of a circle \cite[Chapter~5.1, E.19]{borweinerdelyibk}.
We are not aware of an analogue for weighted polynomials.

We prefer to discuss the choice of the norming subset in  a greater generality. This is done in Section~\ref{normsetsect}. 
The probabilistic estimates on the norms of polynomials are obtained in a more general context as well in Section~\ref{probsect}.
The proof of the Videnskii inequality (Theorem~\ref{ballberntheo}) and Theorem~\ref{locprobesttheo} are given in Section~\ref{pfsect}.
 
\subsection{Norming sets}\label{normsetsect}
 If $\XX$ is a topological space, we denote by $C_0(\XX)$ the space of all continuous real valued functions on $\XX$ vanishing at infinity, equipped with the supremum norm. 

\begin{definition}\label{normnumberdef}
Let $\XX$ be a topological space, $W\subset C_0(\XX)$.
 A set $\C\subset\XX$ is called a \textbf{norming set} for $W$ if there exists a finite number $c>0$ such that
\be\label{normingset}
\sup_{x\in\XX}|f(x)|\le c\sup_{y\in\C}|f(y)|, \qquad f\in W.
\ee
We denote  by $\mathfrak{n}(W, \C)$ the infimum of all the numbers $c$ that work above. 
\end{definition}
\begin{definition}
Let $\mathbb{Y}$ be a metric space with metric $\rho$. If $\epsilon>0$, a subset $V\subseteq \mathbb{Y}$ is called an $\epsilon$-cover of $\mathbb{Y}$ if 
$$
\sup_{f\in \mathbb{Y}}\inf_{g\in V}\rho(f,g) \le \epsilon.
$$
\end{definition}

\begin{prop}\label{normingsetprop}
Let $\XX$ be a topological space, $W$ be a linear subspace of $C_0(\XX)$. \\
{\rm (a)} If there exists a finite norming set $\C$ for $W$ then $W$ is finite dimensional, and $\mathsf{dim}(W)\le |\C|$.\\
{\rm (b)} Let $W$ be  finite dimensional, 
$$
B_W=\{f\in W: \|f\|_{\infty,\XX} = 1\},
$$
and $\{f_1,\cdots,f_N\}$ be a $1/4$-cover for $B_W$. Then there exists a norming set $\C$ for $W$ with $|\C|=N$, and $\mathfrak{n}(\C, W)\le 2$.\\
{\rm (c)} Let $\XX$ be a compact metric space with metric $\rho$, $W$ be a finite dimensional linear subspace of $C_0(\XX)$, and 
there exist $L=L(W)>0$ such that
\be\label{lipschitz}
|f(x)-f(y)|\le L\|f\|_{\infty,\XX}\rho(x,y), \qquad f\in W.
\ee
If $\C$ is a $1/(2L)$-cover for $\XX$, then $\C$ is a norming set for $W$ with $\mathfrak{n}(\C,W)\le 2$.
\end{prop}

\begin{Proof}\ 

If $\C$ is a finite norming subset for $W$ then the mapping $W\to \RR^{|\C|}$ given by $f\mapsto (f(x))_{x\in\C}$ is injective. This proves part (a). 

To prove part (b), let $\{f_1,\cdots, f_N\}$ be a $1/4$-cover for $B_W$. Since each of the functions $f_j\in C_0(\XX)$, there exists $x_j\in\XX$ for which $|f_j(x_j)|=\|f_j\|_{\infty,\XX}$.
We let $\C=\{x_1,\cdots,x_N\}$. If $f\in B_W$, and $\|f-f_j\|_{\infty,\XX} \le 1/4$, then
$$
1=\|f\|_{\infty,\XX}\le \|f-f_j\|_{\infty,\XX}+\|f_j\|_{\infty,\XX}\le 1/4+|f_j(x_j)| \le 1/4+|f_j(x_j)-f(x_j)|+|f(x_j)|\le 1/2+\max_{1\le j\le N}|f(x_j)|.
$$
This proves part (b).

Let the hypothesis of part (c) be satisfied, and $\C$ be a $1/(2L)$ covering for $\XX$. If $f\in W$ and $\|f\|_{\infty,\XX}=|f(x^*)|$, then 
there exists $z\in\C$ with $\rho(x^*,z)\le 1/(2L)$. Then \eref{lipschitz} implies that
$$
\|f\|_{\infty,\XX}\le |f(z)|+|f(x^*)-f(z)|\le \max_{y\in\C}|f(y)|+L\rho(x^*,z)\|f\|_{\infty,\XX} \le \max_{y\in\C}|f(y)| +(1/2)\|f\|_{\infty,\XX}.
$$
This proves part (c).
\end{Proof}

\subsection{Probabilistic estimates}\label{probsect}
We recall Hoeffding's inequality \cite[Appendix~B(3)]{pollard2012bk}:

\begin{prop}\label{hoeffineqprop}
If $Y_1,\cdots,Y_M$ are independent random variables with mean $0$ such that $a_j\le Y_j\le b_j$ for $j=1,\cdots,M$, then for $\eta>0$,
\be\label{hoeffineq}
\mathsf{Prob}\left(\frac{1}{M}\left|\sum_{j=1}^M Y_j\right| \ge \eta\right) \le 2\exp\left(-\frac{M^2\eta^2}{\sum_{j=1}^M (b_j-a_j)^2}\right).
\ee
\end{prop}

Next, we wish to develop a general theorem estimating the probability of a lower bound on the  supremum norm of a random family of functions analogous to \cite[Lemma~6.2]{quadconst}.

\begin{theorem}\label{supnormtheo}
Let $\XX$ be a topological space, $W$ be a linear subspace of $C_0(\XX)$. We assume that there is a finite norming set $\C$ for $W$. Let $(\Omega, \mathcal{B}, \mu)$ be a probability space, and $Z : \Omega\to W$. We assume further that for any $x\in\XX$, $\omega\in\Omega$, $|Z(\omega)(x)|\le R$ for some $R>0$. Then for any $\delta>0$, integer $M\ge 1$, and independent sample $\omega_1,\cdots, \omega_M$, we have
\be\label{genprobest}
\mathsf{Prob}_\mu\left(\sup_{x\in\XX}\left|\frac{1}{M}\sum_{j=1}^M Z(\omega_j)(x)-\mathbb{E}_\mu(Z(\circ)(x))\right|\ge 4\mathfrak{n}(W,\C)R\sqrt{\frac{\log(2|\C|/\delta)}{M}}\right) \le \delta.
\ee
\end{theorem}

\begin{Proof}\ 
Let $x\in\XX$, and $Y_j=Z(\omega_j)(x)-\mathbb{E}_\mu(Z(\circ)(x))$. Then $|Y_j|\le 2R$, and Hoeffding's inequality implies that for any $\eta>0$,
\be\label{pf1eqn1}
\mathsf{Prob}_\mu\left(\left|\frac{1}{M}\sum_{j=1}^M Z(\omega_j)(x)-\mathbb{E}_\mu(Z(\circ)(x))\right|\ge \frac{\eta}{2\mathfrak{n}(W,\C)}\right)\le 2\exp\left(-\frac{M\eta^2}{16\mathfrak{n}(W,\C)^2R^2}\right).
\ee
We observe that since $W$ is a finite dimensional linear space,  the functions 
$$
x\mapsto  \frac{1}{M}\sum_{j=1}^M Z(\omega_j)(x)-\mathbb{E}_\mu(Z(\circ)(x)
$$ 
are in $W$ for all choices of the $\omega_j$'s. 
We now apply \eref{pf1eqn1} with each $y\in \C$ in place of $x$ to conclude that
\begin{eqnarray*}
\lefteqn{\mathsf{Prob}_\mu\left(\sup_{x\in\XX}\left|\frac{1}{M}\sum_{j=1}^M Z(\omega_j)(x)-\mathbb{E}_\mu(Z(\circ)(x))\right|\ge \eta\right)}\\
&\le& \bigcup_{y\in\C} \mathsf{Prob}_\mu\left(\left|\frac{1}{M}\sum_{j=1}^M Z(\omega_j)(y)-\mathbb{E}_\mu(Z(\circ)(y))\right|\ge \frac{\eta}{2\mathfrak{n}(W,\C)}\right)\\
&\le& 2|\C|\exp\left(-\frac{M\eta^2}{16\mathfrak{n}(W,\C)^2R^2}\right).
\end{eqnarray*}
We set the last expression above to $\delta$ and solve for $\eta$ to obtain \eref{genprobest}.
\end{Proof}

The following lemma states the asymptotics in \eref{lambertass} for the Lambert function in \eref{lambertfn} in a form that is easily applicable in the proof of \eref{locprobest} in Theorem~\ref{locprobesttheo}.

 \begin{lemma}\label{lambertlemma}
Let $y, \alpha, \beta, A, B>0$. The solution of the equation
\be\label{lamberteqn}
Ax^\alpha\log(Bx^\beta)=y
\ee
is given by
\be\label{lambertsolution}
x=B^{-1/\beta}\exp\left(\frac{1}{\alpha}W\left(\frac{B^{\alpha/\beta}}{A}\frac{\alpha}{\beta}y\right)\right)\approx \left(\frac{\alpha}{\beta A}\right)^{1/\alpha}\left\{\frac{ y}{\disp\log y +\log\left(\frac{\alpha B^{\alpha/\beta}}{\beta A}\right)}\right\}^{1/\alpha},
\ee
where $\approx$ denotes an asymptotic relationship.
 \end{lemma}

 \begin{Proof}\ 
 We multiply both sides of \eref{lamberteqn} by $\alpha/\beta$, and write  $\disp z=\log\left(B^{\alpha/\beta}x^\alpha\right)$ to obtain
$$
e^z=B^{\alpha/\beta} x^\alpha,\quad ze^z=\frac{\alpha}{\beta}\frac{B^{\alpha/\beta}}{A}y.
$$
The solution of this equation leads to the first expression on the right hand side of \eref{lambertsolution}. The asymptotic expression in \eref{lambertass} leads to the second expression.
 \end{Proof}

\subsection{Proof of Theorem~\ref{locprobesttheo}}\label{mainpfsect}
We wish to obtain a Videnskii inequality for weighted polynomials; i.e., an analogue of \eref{bernineq} for derivatives on a ball. We note that the following bound is much worse than \eref{bernineq}, but is adequate for our purpose. 

\begin{theorem}\label{ballberntheo}
Let $n\ge 1$, $r>c/n^2$, $\x_0+2r\in [-2n, 2n]^q$,  $P\in\Pi_n^q$. Then there exists constant $C>0$ independent of $r$, $n$, and $\x_0$, and depending linearly on $q$ 
such that
\be\label{ballbernineq}
\||\nabla P|\|_{\infty,\mathbb{B}(\x_0,r)} \le  Cre^{1/r}n^4\|P\|_{\infty, \mathbb{B}(\x_0,r)}.
\ee
\end{theorem}

\begin{Proof}\ 
It is sufficient to prove \eref{ballbernineq} for $q=1$; the general case follows by applying the univariate inequality one component at a time.
We will simplify our notation by writing $m=n^2$ in this proof. Let $P\in \Pi_n^1$, and $Q$ be a (univariate) polynomial of degree $<m$ such that $P(x)=Q(x)\exp(-x^2/2)$. Without loss of generality, let
$\|P\|_{\infty, [x_0-r,x_0+r]}=1$.
We consider the function $G :\CC\to \CC$ defined by
\be\label{pf3eqn1}
G(z)=\left\{\frac{z-x_0+\sqrt{(z-x_0)^2-r^2}}{r}\right\}^m\exp\left(-\frac{(z+x_0)\sqrt{(z-x_0)^2-r^2}}{2}\right),
\ee
where the principle branch of the square root is chosen (so that $\sqrt{\zeta}\to \infty$ as $\zeta\to\infty$).
The mapping
\be\label{pf3eqn2} 
 w= \frac{z-x_0+\sqrt{(z-x_0)^2-r^2}}{r}
 \ee
maps the exterior of $[x_0-r,x_0+r]$ to the exterior of the complex unit disc, and obviously, $|w|=1$ if $z\in [x_0-r,x_0+r]$. So, $|G(z)|=1$ if $z\in [x_0-r,x_0+r]$. Hence, the function
$$
F(z)=\frac{Q(z)\exp(-z^2/2)}{G(z)}
$$
is analytic in $\CC\cup\{\infty\}\setminus [x_0-r,x_0+r]$ and $|F(z)|\le 1$ on $[x_0-r,x_0+r]$. Therefore, the maximum modulus principle shows that
\be\label{pf3eqn3}
|P(z)|=|Q(z)\exp(-z^2/2)| \le |G(z)|, \qquad z\in \CC\setminus [x_0-r,x_0+r].
\ee 
We wish to use this bound on the interior of the ellipse $\mathcal{E}$ defined by $|w|=1+1/(rm)$. Since 
$$
z=x_0+\frac{r}{2}(w+1/w),
$$
we calculate that
$$
(z-x_0)^2-r^2=\frac{r^2}{4}\left((w+w^{-1})^2-4\right)=\frac{r^2}{4}(w-w^{-1})^2,
$$
so that
$$
 (z+x_0)\sqrt{(z-x_0)^2-r^2}=x_0r(w-w^{-1})+\frac{r^2}{4}(w^2-w^{-2}).
 $$
Therefore, on the ellipse $\mathcal{E}$ parametrized by $w=(1+(rm)^{-1})\exp(i\theta)$,
$$
\left|\Re e \left((z+x_0)\sqrt{(z-x_0)^2-r^2}\right)\right|= \left|\left(\frac{2x_0}{m}\cos\theta +\frac{4r}{m}\cos(2\theta)\right)(1+\O(1/r^2m^2))\right| \le c\frac{|x_0+2r|}{m}.
$$
Thus, if  $|x_0+2r|\le 2\sqrt{m}$, the estimate \eref{pf3eqn3} shows that for $z$ on and inside $\mathcal{E}$,
\be\label{pf3eqn4}
|P(z)|\le (1+(rm)^{-1})^m\exp\left( c_1/\sqrt{m}\right).
\ee
Now, let $x\in [x_0-r,x_0+r]$. It is easily calculated that the minimum distance between $\mathcal{E}$ and $[x_0-r,x_0+r]$ is at least $2c_2(rm^2)^{-1}$ for a constant $c_2$ independent of $r$ and $m$ as long as $rm\ge c$. 
Hence, the complex disc bounded by the contour $\Gamma : \{\zeta : |\zeta-x|=  c_2(rm^2)^{-1}\}$ is contained in the interior of $\mathcal{E}$. 
On this disc, the bound \eref{pf3eqn4} holds.
Therefore, the Cauchy integral formula  leads to
$$
|P'(x)|=\frac{1}{2\pi}\left|\oint_\Gamma \frac{P(\zeta)}{(\zeta-x)^2}d\zeta\right|\le c_3rm^2e^{1/r}.
$$
Recalling that $m=n^2$, this leads to \eref{ballbernineq} when $q=1$. As explained earlier, this completes the proof.
\end{Proof}

\noindent\textsc{Proof of Theorem~\ref{locprobesttheo}.}\\

Let $\x_0\in\RR^q$, $n\ge 1$, $r\ge c/n^2$ where $c$ is as in Theorem~\ref{ballberntheo}. We apply Theorem~\ref{supnormtheo} with $\XX=\mathbb{B}(\x_0,r)$, $\Pi_n^q$ (restricted to $\XX$) in place of $W$, $\RR^q\times \Omega$ in place of $\Omega$, $\tau$ as in Theorem~\ref{locprobesttheo}.  We take 
$$
Z(z,\epsilon)(\x)=\mathcal{F}(\z,\epsilon)\Phi_n(\x,\z),
$$
 where $(\z,\epsilon)$ is a random sample from $\tau$.  Then the quantity $R$ in Theorem~\ref{supnormtheo} can be chosen to be $c_1\|\mathcal{F}\|_\infty n^q$. Moreover,  \eref{iterexp} shows that,
$$
\mathbb{E}_{\tau}(Z)(\x)=\sigma_n(Ff)(\x).
$$
In view of Theorem~\ref{ballberntheo}, there exists a norming set $\C\subset \mathbb{B}(\x_0,r)$ for $W$ with $|\C|\sim \exp(q/r)(n^4 r^2)^q$ and $\mathfrak{n}(\C, W)\le 2$.
Theorem~\ref{supnormtheo} used with $\delta(r/n)^q$ in place of $\delta$ shows that
\be\label{pf4eqn1}
\mathsf{Prob}_{\mu\times\tau}\left(\left\|\frac{1}{M}\sum_{j=1}^M\mathcal{F}(\y_j, \epsilon_j)\Phi_n(\circ,\y_j)-\sigma_n(Ff)\right\|_{\infty,\mathbb{B}(\x_0,r)}
\!\!\!\ge c_1\|\mathcal{F}\|_\infty n^q
\sqrt{\frac{\log(c_3r^{q}\exp(q/r)n^{5q}/\delta)}{M}}\right)\le \delta(r/n)^q.
\ee
Since $Ff\in W_{\infty,\gamma}(\x_0)$, Theorem~\ref{localchartheo} shows that
\be\label{pf4eqn2}
\|Ff-\sigma_n(Ff)\|_{\infty,\mathbb{B}(\x_0,r)} \le n^{-\gamma}\|Ff\|_{\infty,\gamma,\x_0,r}.
\ee
Therefore, \eref{pf4eqn1} leads to \eref{masterprobest}.

To prove \eref{locprobest}, we  solve for $n$ the equation
$$
n^{2q+2\gamma}\log(Bn^{5q})=M,
$$
where $B$ is defined in the statement of Theorem~\ref{locprobesttheo}.
 Lemma~\ref{lambertlemma}, used with $A=1$, $\alpha=2q+2\gamma$, $\beta=5q$ shows that the solution is given by \eref{nmdef}. Therefore, with this choice of $n$,
  \eref{masterprobest} implies \eref{locprobest}.
 \qed


\end{document}